\documentclass[lettersize,journal]{IEEEtran}
\usepackage{amsmath,amsfonts}
\usepackage{algorithmic}
\usepackage{algorithm}
\usepackage{array}
\usepackage[caption=false,font=normalsize,labelfont=sf,textfont=sf]{subfig}
\usepackage{textcomp}
\usepackage{stfloats}
\usepackage{url}
\usepackage{verbatim}
\usepackage{graphicx}
\usepackage{cite}

\usepackage{adjustbox}
\usepackage{booktabs}
\usepackage{tikz}
\usepackage{multirow}
\usepackage{orcidlink}

\usepackage[T1]{fontenc}
\usepackage{textcomp}

\newcommand{\dq}[1]{\textquotedbl{}#1\textquotedbl{}}

\begin{document}

\title{DAGLFNet: Deep Feature Attention Guided Global and Local Feature Fusion for Pseudo-Image Point Cloud Segmentation}

\author{Chuang Chen\orcidlink{0009-0008-0689-3985}, Yi Lin\orcidlink{0000-0002-7194-5023}, Bo Wang\orcidlink{0009-0007-3394-2642}, Jing Hu\orcidlink{0000-0003-0921-0592}, Xi Wu\orcidlink{0000-0002-7659-1631} and Wenyi Ge\orcidlink{0009-0006-2731-6074}
\thanks{This work was supported in part by the \dq{Juyuan Xingchuan} Project of Central Universities and Research Institutes in Sichuan, High-Resolution Multi-Wavelength Lidar System and Large-Scale Industry Application (2024ZHCG0190), the Key R\&D Project of the Sichuan Provincial Department of Science and Technology-Research on Three-Dimensional Multi-Resolution Intelligent Map Construction Technology (2024YFG0009), the Intelligent Identification and Assessment for Disaster Scenes: Key Technology Research and Application Demonstration (2025YFNH0008), Project of the Sichuan Provincial Department of Science and Technology-Application and Demonstration of Intelligent Fusion Processing of Laser Imaging Radar Data (2024ZHCG0176). ($Corresponding \ author: Wenyi\ Ge.$)}
\thanks{Chuang Chen is with the National Key Laboratory of Fundamental Science on Synthetic Vision, Sichuan University, Chengdu, 610065, China; College of Computer Science, Chengdu University of Information Technology, Chengdu, 610225, China. (e-mail: 3230609001@stu.cuit.edu.cn)}

\thanks{Yi Lin is with the College of Computer Science, Sichuan University, Chengdu, 610065, China. (e-mail: yilin@scu.edu.cn)}

\thanks{ Bo Wang, Jing Hu, Xi Wu, and Wenyi Ge are with the College of Computer Science, Chengdu University of Information Technology, Chengdu, 610225, China. (e-mail: wsunrise2022@163.com, lesley123@cuit.edu.cn, wuxi@cuit.edu.cn, gewenyi15@cuit.edu.cn)}

\thanks{Code will be available on: \url{https://github.com/wtxjbl-tp/DAGLFNet} }

}

\markboth{Journal of \LaTeX\ Class Files,~Vol.~14, No.~8, August~2021}%
{Shell \MakeLowercase{\textit{et al.}}: A Sample Article Using IEEEtran.cls for IEEE Journals}


\maketitle

\begin{abstract}
Environmental perception systems are crucial for high-precision mapping and autonomous navigation, with LiDAR serving as a core sensor that provides accurate 3D point cloud data. In recent years, numerous pseudo-image-based representation methods have emerged to balance efficiency and performance by fusing 3D point clouds with 2D grids. However, pseudo-image representations struggle to preserve distance-dependent 3D structural characteristics, while feature interference from densely distributed near-range points weakens the discriminability of sparse far-range features. We present DAGLFNet, a pseudo-image-based semantic segmentation framework designed to extract discriminative features. DAGLFNet incorporates three key components: first, a Global-Local Feature Fusion Encoding (GL-FFE) module to enhance intra-set local feature correlation and capture global contextual information; second, a Multi-Branch Feature Extraction (MB-FE) network to capture richer neighborhood information and improve the discriminability of contour features; and third, a Feature Fusion via Depth Feature-guided Attention (FFDFA) mechanism to refine cross-channel feature fusion precision. Experimental evaluations demonstrate that DAGLFNet achieves mean intersection-over-union (mIoU) scores of 69.1\% and 78.4\% on the validation sets of SemanticKITTI and nuScenes, respectively. Additional experiments on the ScanNet dataset evaluate its applicability to indoor environments, suggesting that the proposed framework can be adapted to both outdoor and indoor semantic segmentation settings.
\end{abstract}


\section{Introduction}
\IEEEPARstart{S}{emantic} segmentation has become a cornerstone technology for 3D environmental perception, enabling dense semantic annotation and feature learning from LiDAR point clouds \cite{Domain_Adaptive}, \cite{PrimePSegter}, \cite{wu2024prospective}. By providing structured interpretations of complex environments, this capability is fundamental for applications such as robotics and autonomous driving. Modern LiDAR sensors capture tens of millions of points per second, thereby allowing an unprecedentedly detailed representation of the surrounding 3D structure \cite{behley2019semantickitti}, \cite{Panoptic-nuScenes}. However, the central challenge lies in developing effective strategies to process inherently unstructured and unordered point cloud data to extract discriminative features for reliable perception.

Existing LiDAR semantic segmentation methods rely on different point cloud representations, including point-based \cite{hu2020randla}, \cite{qi2017pointnet}, voxel-based \cite{octnet}, \cite{SPVNAS-V1}, \cite{zhou2020cylinder3d}, multi-modal fusion \cite{liu2023uniseg}, \cite{Multi-modal}, and range-based paradigms \cite{milioto2019rangenet++}, \cite{cheng2022cenet}, \cite{zhao2021fidnet}, \cite{RAFDet}. Although point-based \cite{qi2017pointnet} and voxel-based \cite{zhou2020cylinder3d} methods preserve spatial geometry, they often require costly neighborhood searches or fine-grained 3D discretization. Multi-modal methods \cite{liu2023uniseg} improve accuracy by combining LiDAR geometry with image semantics, but incur additional costs from feature extraction, spatial alignment, and cross-modal fusion. In contrast, range-based methods \cite{milioto2019rangenet++}, \cite{kong2023rethinking} are becoming a mainstream approach for LiDAR semantic segmentation by projecting 3D point clouds into structured 2D range images, enabling efficient feature extraction with mature 2D convolutional architectures while maintaining competitive segmentation performance. However, converting 3D point clouds into 2D range images inevitably introduces coordinate conflicts of geometric relationships \cite{xu2023frnet}. Specifically, multiple 3D points may be mapped to the same 2D pixel coordinate. Since each range-image pixel stores only one feature representation, points sharing identical projected coordinates are collapsed into a single feature location, causing part of the point-wise information to be overwritten. Such feature compression distorts the original 3D geometric distribution and makes points belonging to objects at different depths difficult to distinguish in the range image representation. Consequently, fine-grained point-wise geometric relationships are weakened, potentially leading to semantic confusion.

\begin{figure*}[htbp]
    \centering
    \captionsetup[subfloat]{font=scriptsize,labelfont=sf,textfont=sf}

    \subfloat[2D image feature representation.\label{fig:question}]{
        \includegraphics[width=1\linewidth]{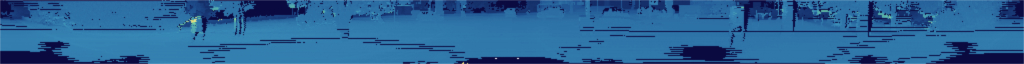}
    }
    \vspace{-0.7em}
    \subfloat[Semantic segmentation result.\label{fig:result}]{
        \includegraphics[width=1\linewidth]{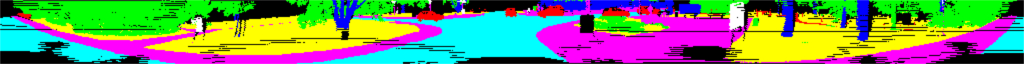}
    }

    \caption{(a) Visualization of encoded 2D image features, and (b) corresponding semantic segmentation result demonstrating the classification performance.}
    \label{fig:question_and_result}
\end{figure*}

\begin{figure}[!htbp]
  \centering
    \captionsetup[subfloat]{font=scriptsize}
  \subfloat[\scriptsize Original LiDAR scan. \label{fig:original_point}]{
      \includegraphics[width=0.45\linewidth]{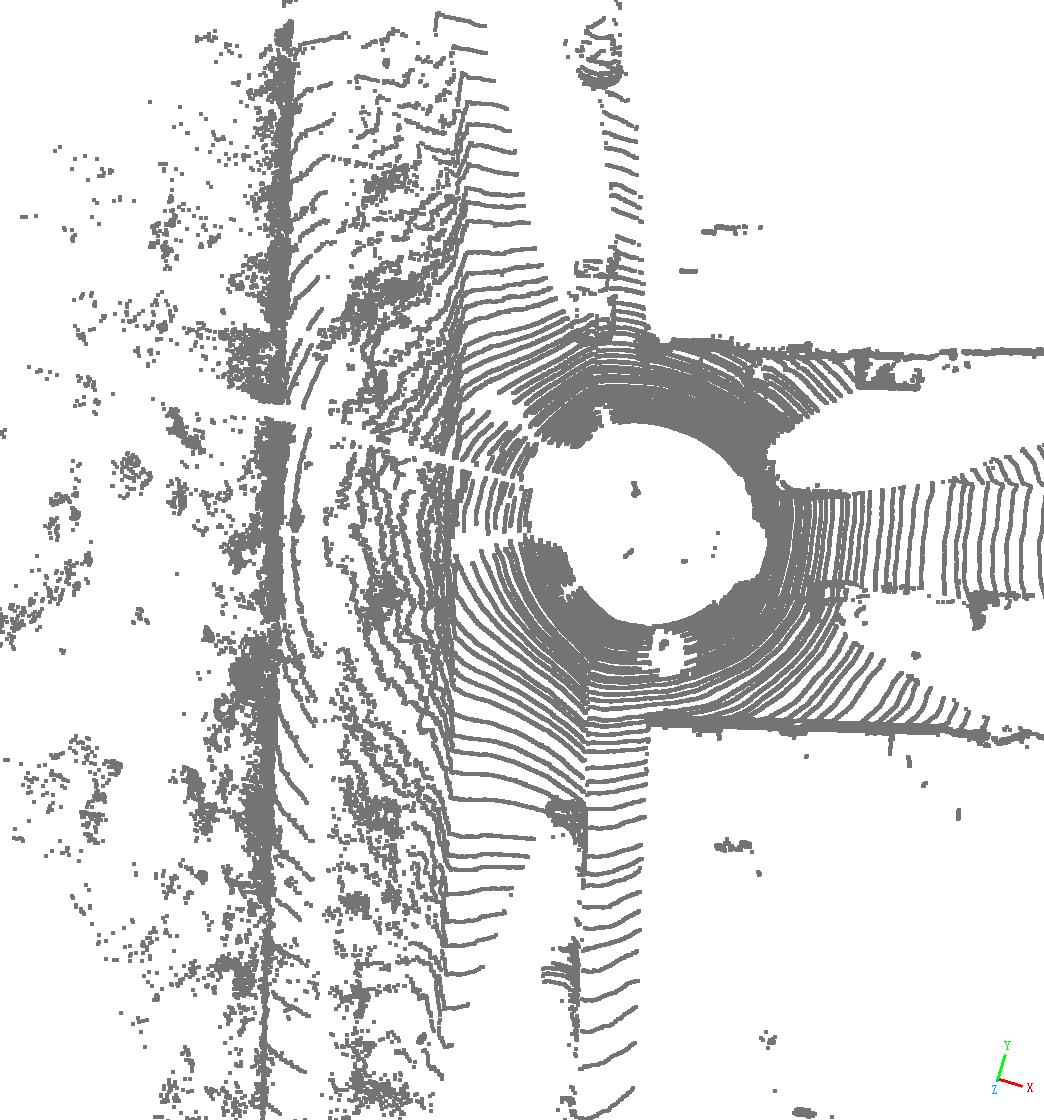}
  }
  \hspace{0.025\linewidth}
  \subfloat[\scriptsize Augmented LiDAR scan. \label{fig:data_argument}]{
      \includegraphics[width=0.45\linewidth]{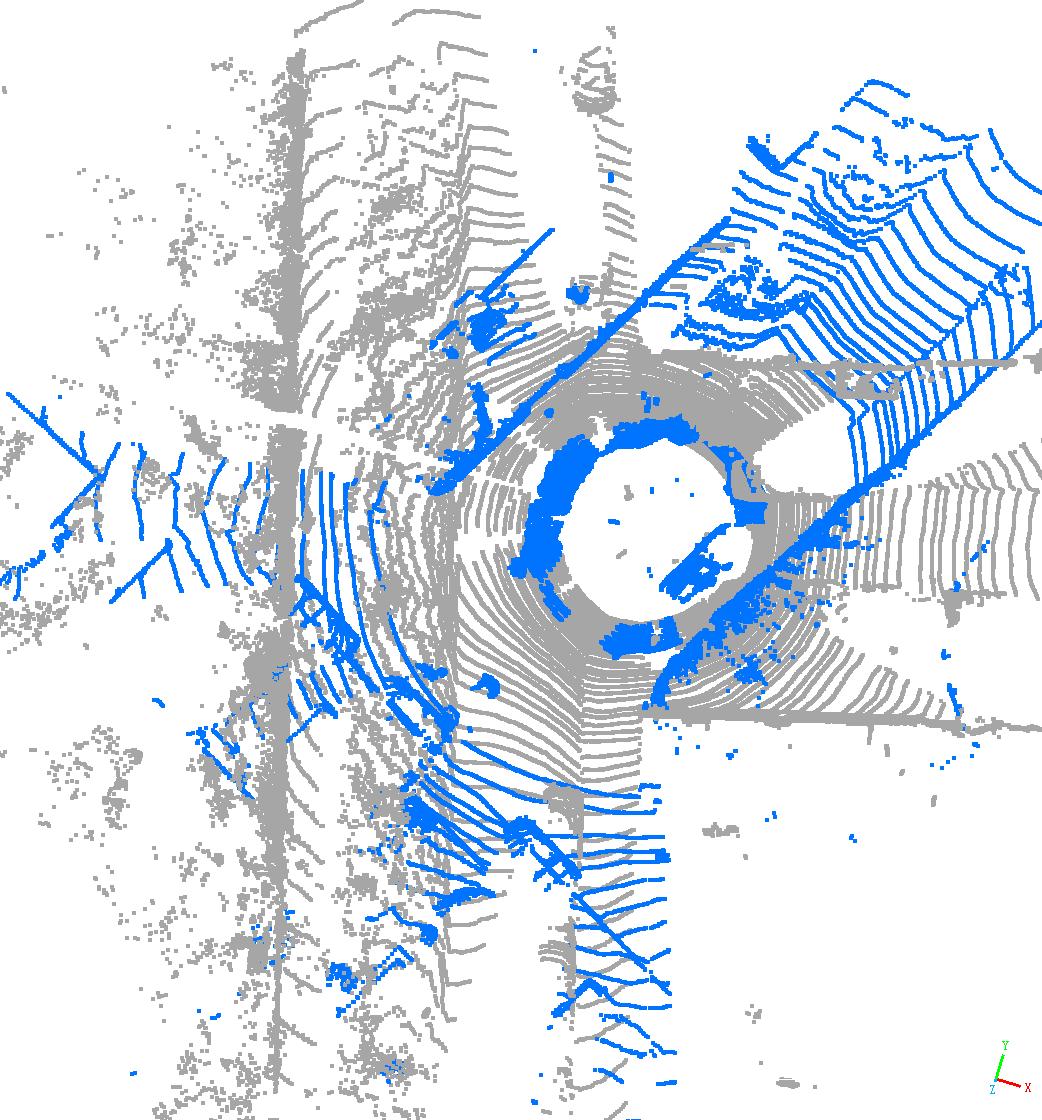}
  }
\captionsetup[sub]{font=scriptsize}
  \caption{
    Illustration of data augmentation, with the original point cloud shown in gray and the inserted points highlighted in blue.
  }
  \label{fig:data_argument_case}
\end{figure}

A common strategy for alleviating projection-induced information loss is to organize point-wise 3D geometric features into a regular 2D grid and process them using image-based feature extraction networks. Derived from the conventional range-based paradigm, this type of representation is commonly referred to as a pseudo-image representation \cite{xu2023frnet}, \cite{FARVNet}. The preservation of additional 3D structural cues within the 2D projected domain helps reduce the information loss and geometric distortion caused by range-view projection. However, due to the lack of explicit distinction between near-range and far-range point features, these methods tend to capture features dominated by locally dense points, with limited ability to represent the complete structural patterns of far-range regions. Therefore, instead of effectively modeling the distance-dependent structural characteristics of point clouds, they are susceptible to interference from near-range point features, which weakens the representation of relatively subtle structural information in far-range regions, resulting in compromised feature discrimination and further deteriorating the final segmentation performance. In addition, as illustrated in Fig.~\ref{fig:question_and_result}, increasing sensing range reduces the LiDAR sampling density, resulting in fewer valid pixels and increasingly incomplete geometric structures and object boundaries in range images. Range-dependent degradation of projected representations limits the extraction of discriminative contextual information for distant objects, while errors generated in the 2D domain are transferred to point-wise predictions during back-projection, further compromising far-range segmentation accuracy. Data augmentation is widely used to enrich training samples and improve model generalization. However, as shown in Fig.~\ref{fig:data_argument_case}, scene-mixing strategies may alter local geometric relationships by inserting points from other scans. Consequently, methods that rely primarily on local feature modeling struggle to extract reliable structural information, thereby weakening feature discriminability.

To address the aforementioned issues, we propose DAGLFNet, a novel network architecture for pseudo-image-based LiDAR semantic segmentation that aims to enhance the representation of far-range point clouds. Specifically, to compensate for the weakened point-wise geometric relationships caused by range-view projection, we design a Global-Local Feature Fusion Encoding (GL-FFE) module that explicitly encodes local geometric interactions among neighboring points while propagating contextual information across different projected regions, allowing incomplete local structural information to be complemented by global contextual cues. To improve object-boundary representation in range images, we develop a Multi-Branch Feature Extraction (MB-FE) network that employs parallel branches with different receptive fields to capture complementary structural information at multiple spatial scales, allowing fine-grained local details and large-scale object contours to be jointly modeled. In addition, to alleviate feature interference between near-range and far-range regions, we introduce a Feature Fusion via Depth Feature-guided Attention (FFDFA) mechanism that leverages depth information to guide feature fusion, enabling the network to distinguish features at different sensing distances and enhance the representation of distant regions. Extensive experiments on the outdoor SemanticKITTI \cite{behley2019semantickitti} and nuScenes \cite{Panoptic-nuScenes} datasets, together with additional evaluations on the indoor ScanNet \cite{dai2017scannet} dataset, demonstrate the effectiveness and generalization capability of the proposed network architecture across diverse scenarios.
\\
In summary, our main contributions are as follows:

\begin{enumerate}
\item We propose DAGLFNet, a novel pseudo-image-based LiDAR semantic segmentation network that explicitly addresses geometric information loss, ambiguous range-view representation, and feature interference across different sensing distances, enabling more discriminative feature learning for distant point clouds.
\item A Global-Local Feature Fusion Encoding (GL-FFE) module and a Multi-Branch Feature Extraction (MB-FE) network are developed to enhance geometric representation, contextual interaction, and the modeling of associations between global information and local features.
\item A Feature Fusion via Depth Feature-guided Attention (FFDFA) mechanism is proposed to suppress feature interference between near-range and distant regions through depth-aware adaptive fusion.
\item Extensive experiments on two widely adopted LiDAR segmentation benchmarks demonstrate the effectiveness of DAGLFNet. Specifically, our method achieves competitive mIoU scores of 69.1\% and 78.4\% on the SemanticKITTI and nuScenes validation sets, respectively. Additional experiments are also conducted on the ScanNet dataset to evaluate performance in indoor scenes. The comprehensive analysis further confirms the validity of the research motivation.
\end{enumerate}

\section{Related Work}

\subsection{Point Cloud Feature Encoding}

Semantic segmentation of LiDAR point clouds assigns semantic labels to individual points in a scene, transforming unstructured data into structured environmental understanding \cite{Semantic}, \cite{Region-enhanced}. Encoding features from raw point clouds is crucial for extracting three-dimensional spatial information by characterizing the spatial distribution and geometric relationships among points through different data organization schemes. Existing methods mainly process raw point clouds through point modeling, voxel-based representations, and range-based representations.

Point-based methods directly learn features from point coordinates and attributes. PointNet \cite{qi2017pointnet} pioneered the use of multilayer perceptrons (MLPs) \cite{mlp} and symmetric pooling operations to obtain permutation-invariant representations of point sets. Subsequent methods further introduced point convolutions \cite{ShellNet}, \cite{PAConv}, graph convolutions \cite{PU-GCN}, \cite{AGConv}, and attention mechanisms \cite{VANIAN2022277} to capture local geometric structures through neighborhood feature aggregation. Voxel-based methods instead discretize point clouds into regular 3D grids. OctNet \cite{octnet} hierarchically organizes occupied regions, MinkowskiNet \cite{MinkNet} models sparse voxels using Minkowski convolutions, and Cylinder3D \cite{zhou2020cylinder3d} combines cylindrical partitioning with asymmetric 3D convolutions. Range-based methods organize raw points according to the angular sampling pattern of LiDAR sensors. RangeNet++ \cite{milioto2019rangenet++} encodes point coordinates, range, and reflectance into spherical grids, while FRNet \cite{xu2023frnet} groups points sharing the same angular interval into frustums and aggregates their attributes into structured representations. However, the above methods are primarily designed around the spatial organization of raw point clouds and overlook the influence of data augmentation on the geometric relationships established during feature encoding. In particular, scene-mixing augmentation strategies enrich the training data distribution and improve model generalization, but may also alter local spatial structures and introduce geometric relationships inconsistent with the surrounding scene. In this work, we propose GL-FFE to encode local point-cloud geometry and propagate contextual information across projected regions, thereby mitigating local geometric inconsistencies introduced by data augmentation.

\begin{figure*}[!b]
    \centering
    \includegraphics[width=\linewidth]{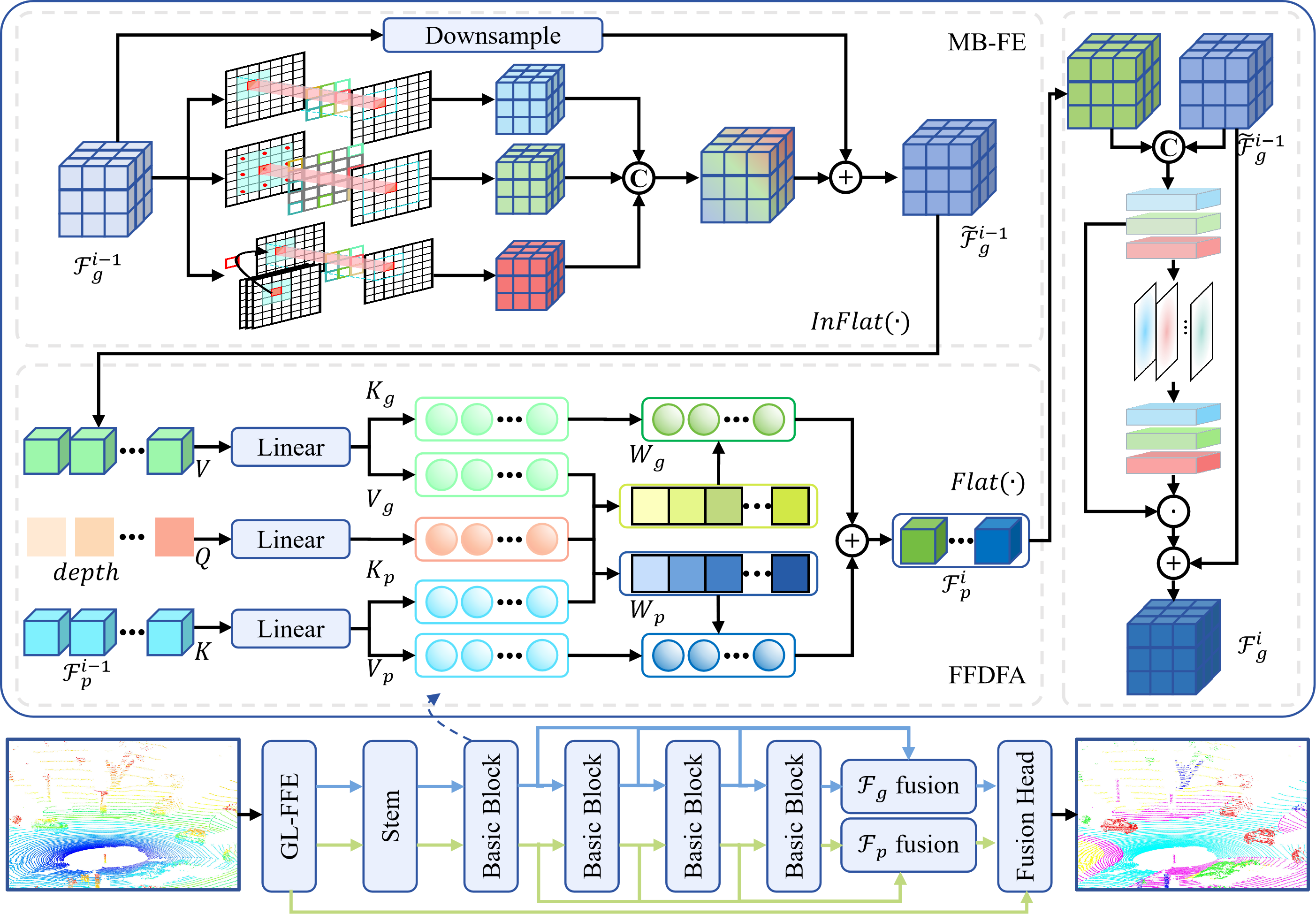}
    \caption{The proposed DAGLFNet framework consists of key components such as GL-FFE, MB-FE, FFDFA, and the Fusion Head, which are responsible for contextual and geometric feature extraction, boundary enhancement, multi-scale feature integration, and final prediction respectively. Multiple stacked DAGLFNet units continuously learn complex hierarchical features from the point cloud, with the Fusion Head combining point-level and group-level features to predict the final output.}
    \label{flowchart}
\end{figure*}

\subsection{Projection-Based Segmentation}

\label{2D Representation for Semantic Segmentation}
To achieve a balance between performance and efficiency in LiDAR point cloud semantic segmentation, researchers have proposed range image-based semantic segmentation methods \cite{milioto2019rangenet++},\cite{kong2023rethinking}, \cite{xu2023frnet}, \cite{FARVNet}. RangeNet++ \cite{milioto2019rangenet++} employs a convolutional encoder and decoder to process spherical range images. SalsaNet \cite{aksoy2020salsanet} and CENet \cite{cheng2022cenet} enhance projected image features by enlarging the receptive field and aggregating multiscale contextual information, while RangeFormer \cite{kong2023rethinking} captures dependencies over long distances through hierarchical attention. Building on these advances, subsequent methods incorporate raw point cloud information to complement projected representations. FRNet \cite{xu2023frnet} retains the original points within each frustum and hierarchically propagates information between point features and projected image features. FARVNet \cite{FARVNet} further combines geometric and reflectivity reconstruction with adaptive multiscale fusion to enhance projected representations and their interaction with point information. However, the aforementioned methods generally adopt a uniform scheme for projected feature extraction and interact with point features through projection correspondences, making it difficult to accommodate nonuniform feature distributions across different sensing ranges. In this work, MB-FE and FFDFA are introduced to improve the encoding of projected features and their interaction with point features, respectively. At each encoding stage, MB-FE models projected features through parallel branches with different receptive fields and integrates their responses before the residual update, thereby improving the representation of nonuniform features across sensing distances. FFDFA then uses depth information to regulate the relative contributions of the MB-FE output and point features during fusion. The updated point features are subsequently projected back into the image branch, allowing the fused information to participate continuously in feature refinement at subsequent stages.

\section{Methodology}

Building on the limitations identified in Section \ref{2D Representation for Semantic Segmentation}, we propose a unified framework that jointly optimizes 2D projection and 3D structural fidelity. Our framework consists of four complementary components: a GL-FFE module for capturing contextual and local geometric dependencies, a MB-FE network for enhancing boundary representations, a FFDFA strategy for precise integration of multi-scale information, and a fusion head to integrate low-level and high-level semantic information.

\subsection{Problem Definition}

Given a point cloud $P = \{p_1, p_2, \ldots, p_N\}$, where each point $p_j = (x_j, y_j, z_j, I_j)$ consists of its 3D coordinates $(x_j, y_j, z_j)$ and reflectivity $I_j$ captured by the LiDAR sensor, our network aims to predict a semantic label $l_j \in \mathcal{C}$ for each point $p_j$ through the network architecture $\mathcal{G}$. Formally, the network can be defined as:

\begin{equation}
    L = \mathcal{G}(P,\theta)
\end{equation}

where $\theta$ represents the learnable parameters within the network. DAGLFNet consists of multiple stacked basicblocks that continuously update point cloud features and image features through hierarchical interaction and fusion to alleviate information loss caused by point cloud projection, as shown in Fig.~\ref{flowchart}. Initial point cloud feature extraction directly influences the quality of subsequent feature interaction and fusion. MB-FE enhances image feature representation to alleviate feature ambiguity in range images. The FFDFA mechanism leverages depth-aware weighting to perform adaptive cross-channel feature fusion, alleviating feature interference between near-range and distant regions. The fusion head aggregates multi-level point cloud features and aligned image features to generate final semantic predictions.

\subsection{Feature Encoder}

\begin{figure*}[!b]
    \centering
    \includegraphics[width=\linewidth]{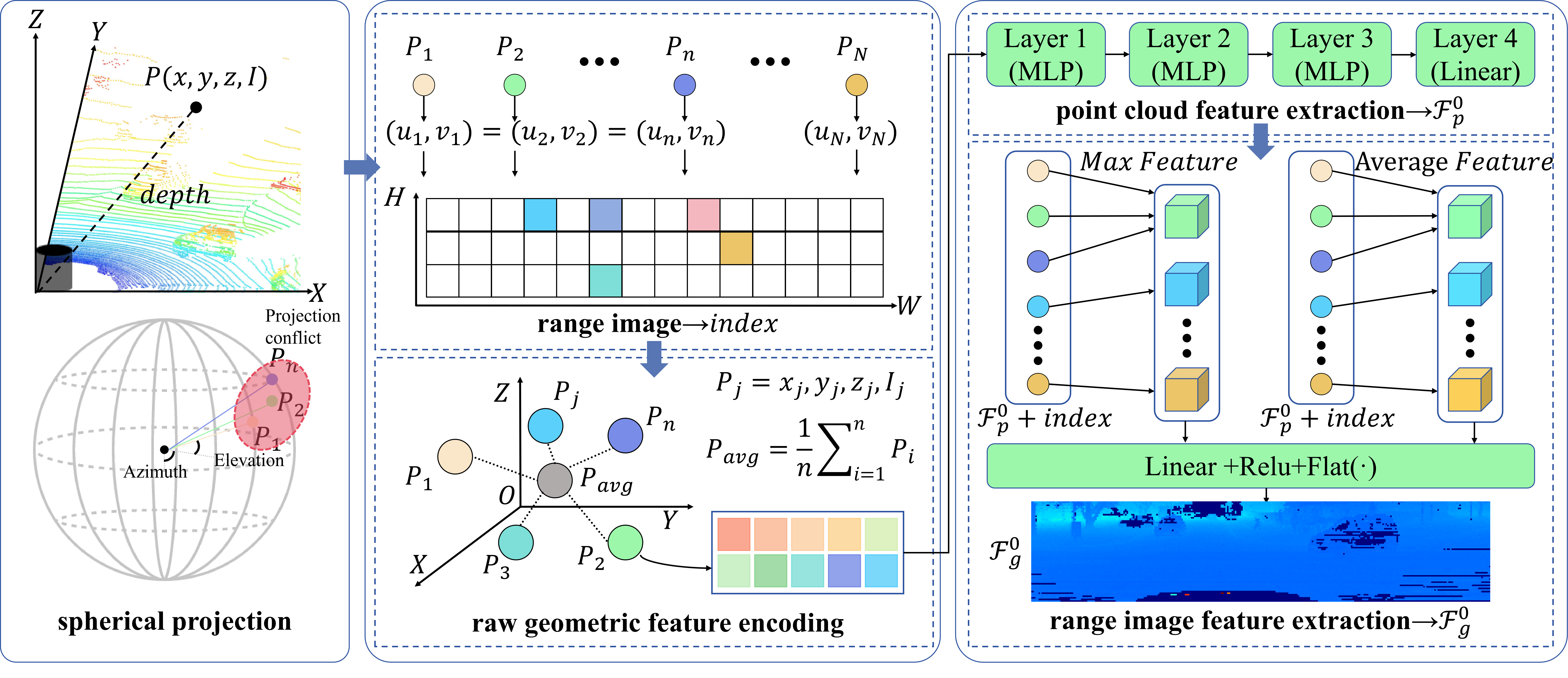}
    \caption{Overview of the proposed GL-FFE module for global-local feature interaction and fusion.
}
    \label{GL-FFE}
\end{figure*}

Fig.~\ref{GL-FFE} illustrates the architecture of the GL-FFE module. Given an input point cloud, the encoder extracts local and global features, which are then projected onto a 2D feature map. Unlike conventional voxelization or regional partitioning approaches \cite{zhou2020cylinder3d,milioto2019rangenet++}, a grouping strategy based on laser beam characteristics is employed, where all laser beams are aligned to a common origin to partition the point cloud into $M$ groups. $P={\{\mathcal{P}_{1},\mathcal{P}_{2},...,\mathcal{P}_{M}\} }$, as follows:

\begin{equation}
    \left\{\begin{matrix}
      r_{i} = \sqrt{x_{i}^{2}+y_{i}^{2}+(z_{i}-h_{l})^{2}}\\
    {v}_{i}= (1-(\text{asin}((z_{i}-h_{l})r_{i}^{-1})+\varphi _{down})\varphi^{-1})H \\
    u_i = \left(1 - (\text{atan}(y_i, x_i) + \pi)(2\pi)^{-1} \right) W
    
\end{matrix}\right.
\label{Hough Voting}
\end{equation}

where $h_l$ denotes the height correction of the $l$-th laser beam; $\varphi = |\varphi_{\text{up}}| + |\varphi_{\text{down}}|$ represents the vertical field-of-view (FoV) of the sensor, 
with $\varphi_{\text{up}}$ and $\varphi_{\text{down}}$ being the inclination angles in the upward and downward directions, respectively; $(v_{i},u_{i})$ denotes the grid coordinate of a point in the range image with a resolution of $H \times W$. After beam-wise height correction, the projected coordinates obtained from Eq.~\ref{Hough Voting} are quantized into integer row and column indices. Each occupied grid cell defines a point group, and $M$ denotes the number of occupied cells. When multiple points fall into the same cell, they are retained within the same group rather than filtered according to depth.

To comprehensively capture geometric and contextual information within each group, we compute the virtual center $p_{avg}$ by averaging all points in the group. The initial feature representation of each point, denoted as $\bar{p}_{j} \in \mathbb{R}^{1\times10}$, is constructed by integrating multiple complementary information sources, including absolute 3D coordinates, reflectance intensity, depth (euclidean distance to the sensor), and relative offsets to the group center $p_{avg}$. The feature vector for each point is defined as:

\begin{equation}
    \bar{p}_{j}=([x_j,y_j,z_j,I_j,depth_j,p_{j}-p_{avg},depth_{p_{j}-p_{avg}}])
\end{equation}

Collectively, the features of all points form the point cloud feature matrix $\bar{P} =\{\bar{p}_{1},\bar{p}_{2},...,\bar{p}_{N}\}\in {\mathbb{R}^{N\times 10}}$. Subsequently, MLPs \cite{mlp} extract discriminative point cloud features.  Each point feature $\bar{p}_j$ is first encoded by MLPs, yielding the initial per-point feature representation $\mathcal{F}_p^0 \in \mathbb{R}^{N \times C}$. Although the point-level features \(\mathcal{F}_p^0\) encode the individual attributes of each point and its relative geometric information with respect to the corresponding group center, they do not explicitly model the geometric relationships among points within each group. To address this limitation, the point features within each group \(\mathcal{P}_m\), where \(m=1,\ldots,M\), are aggregated to obtain compact group-level representations. Specifically, point-level features are first aggregated using max pooling and average pooling to capture complementary structural information, after which the pooled features are concatenated and passed through a linear layer followed by ReLU activation to produce the group-level features $\mathcal{F}_{g}^{0} \in \mathbb{R}^{H \times W \times C}$:

\begin{equation}
\left\{
\begin{aligned}
\mathcal{F}_{p}^{0} &= \text{MLPs}(\bar{P}), \\
F_{\text{cat}} &= [\,\text{MAX}(\mathcal{F}_{p}^{0});\, \text{AVE}(\mathcal{F}_{p}^{0})\,], \\
\mathcal{F}_{g}^{0} &= \text{Flat}\big(\text{ReLU}(\text{Linear}(F_{\text{cat}}))\big).
\end{aligned}
\right.
\end{equation}

where $\text{Flat}(\cdot): \mathbb{R}^{M \times C} \to \mathbb{R}^{H \times W \times C}$ denotes the function that maps point features onto the range image plane with resolution $(H, W)$. $\mathcal{F}_{p}^{0}$ preserves individual point characteristics, while $\mathcal{F}_{g}^{0}$ incorporates global contextual understanding through aggregated group-level features, forming a complementary representation.

\subsection{Image Feature Extraction}
\label{Image Feature Extration}

An MB-FE module processes projected group-level features $\mathcal{F}_{g}^{i-1}$ to address sparsity and blurred boundaries in pseudo-image representations. Following prior work \cite{xu2023frnet, Multi-Branch, kong2023rethinking}, we construct the backbone using multiple convolutional blocks. Each block serves as a core multi-branch unit capturing features at diverse receptive fields. Branch 1) employs a standard $3 \times 3$ convolution to extract local features; Branch 2) uses a dilated $3 \times 3$ convolution (dilation=2) to enlarge the receptive field and capture richer contextual information; Branch 3) focuses on edge enhancement by first reducing the channel dimension with a $1 \times 1$ convolution, followed by a $3 \times 3$ convolution. Subsequently, the three branch outputs undergo channel-wise concatenation and feature fusion through a $1 \times 1$ convolution. A residual connection then integrates the fused features with the input, followed by an activation function. This process can be expressed as:

\begin{equation}
\left\{
\begin{aligned}
{F}_{1} &= \sigma\big(\text{Conv}_{3\times3}(\mathcal{F}_{g}^{i-1})\big), \\[3pt]
{F}_{2} &= \sigma\big(\text{Conv}_{3\times3}^{d=2}(\mathcal{F}_{g}^{i-1})\big), \\[3pt]
{F}_{3} &= \sigma\big(\text{Conv}_{3\times3}\big(\sigma(\text{Conv}_{1\times1}(\mathcal{F}_{g}^{i-1}))\big)\big), \\[3pt]
{F}_{\text{temp}} &= \text{Conv}_{1\times1}([{F}_{1}, {F}_{2}, {F}_{3}]), \\[3pt]
{\tilde{F}_{g}^{i-1}} &= \sigma\big({F}_{\text{temp}} + \mathcal{F}_{g}^{i-1}\big)
\end{aligned}
\right.
\end{equation}

Here, $\text{Conv}_{k\times k}$ denotes a 
$k \times k$ convolution, $\text{Conv}_{3\times 3}^{d=2}$ denotes a dilated convolution with a dilation rate of 2, and $\sigma (\cdot)$ represents the normalization and activation applied after the convolution.

\subsection{Feature Update Module}
\label{Feature Update Module}

The input to the feature update module consists of point-level features $\mathcal{F}_{p}^{i-1}$ and group-level features $\tilde{F}_{g}^{i-1}$, where $\tilde{F}_{g}^{i-1}$ is the output of the MB-FE network from the current stage (Section \ref{Image Feature Extration}). To preserve feature consistency across representations, we incorporate local group-level features into point-level representations through a depth-guided fusion mechanism, which maintains geometric relationships and enhances feature discriminability. The group-level feature representation $\tilde{F}_{g}^{i-1}$ first undergoes projection into the corresponding point-level feature space $\tilde{F}_{p}^{i-1}$ according to the projection index using the operator $\text{InFlat}(\cdot)$. To enable effective cross-feature interaction in the attention mechanism, linear transformations operate on both the projected features $\tilde{F}_{p}^{i-1}$ and the original point-level features ${F}_{p}^{i-1}$, mapping them into a unified latent space. Unlike conventional attention mechanisms \cite{LIU2024104924,HiDAnet}, FFDFA introduces depth information as a dynamic modulation factor to adaptively refine feature weighting. The depth values, denoted by $depth$, are linearly transformed to generate the query $Q$. Depth-guided adjustment alleviates feature interference caused by the mixing of near-range and far-range point clouds within the same representation space during feature fusion:

\begin{equation}
\left\{
\begin{aligned}
V_{g},K_{g} &= \text{Linear}(\text{InFlat}(\tilde{F}_{g}^{i-1}))
\\
V_{p} ,K_{p} &= \text{Linear}({F}_{p}^{i-1})
\\
Q &= \text{Linear}(depth)
\end{aligned}
\right.
\end{equation}

The $\tilde{F}_{g}^{i-1}$ is projected into key and value representations, $K_{\text{g}}$ and $V_{\text{g}}$, respectively, while the $F_{p}^{i-1}$ are similarly transformed into $K_{\text{p}}$ and $V_{\text{p}}$. The attention weights $W_{\text{g}}$ and $W_{\text{p}}$ are computed by measuring the similarity between the depth query $Q$ and the corresponding keys $K_{\text{g}}$ and $K_{\text{p}}$, followed by softmax normalization. These weights are then applied to their respective value matrices and aggregated to obtain the fused representation $F_{{fuse\_p}}$. Finally, $F_{{fuse\_p}}$ undergoes a linear transformation to produce the output feature $\mathcal{F}_{p}^{i}$:

\begin{equation}
\left\{
\begin{aligned}
W_{g} &= \text{softmax}(Q \odot K_{g})\\
W_{p} &= \text{softmax}(Q \odot K_{p})\\
F_{fuse\_p} &= W_{g}\odot V_{g} + W_{p}\odot V_{p}\\
 \mathcal{F}_{p}^{i} &= \text{Linear}(F_{fuse\_p})
\end{aligned}
\right.
\end{equation}

Given the inherent ambiguity of image features, conventional convolution operations may further weaken feature discriminability, thereby compromising the robustness of the learned representations. To counteract the degradation of image feature discriminability introduced by convolutional operations, the updated point-level features $\mathcal{F}_{p}^{i}$ are re-projected onto the image space by $\text{Flat}(\cdot)$, concatenated with $\mathcal{F}_{g}^{i-1}$, and fused through convolutional processing:

\begin{equation}
    \tilde{F}_{fuse\_g} = \sigma (\text{Conv}_{3\times 3 }([\text{Flat}(\mathcal{F}_{p}^{i});\tilde{\mathcal{F}}_{g}^{i-1}]))
\end{equation}

To alleviate feature degradation caused by multiple convolutional operations, an attention map is generated from the fused group-level feature \(\tilde{\mathcal{F}}_{fuse\_g}\) to adaptively recalibrate its responses. The enhanced feature is then added to the input group-level feature \(\mathcal{F}_g^{i-1}\) through a residual connection:

\begin{equation}
     \mathcal{F}_{g}^{i} = \mathcal{F}_{g}^{i-1} +\phi(\text{Conv}_{3 \times 3}(\sigma(\text{Conv}_{3 \times 3}(\tilde{F}_{fuse\_g}))))\odot \tilde{F}_{fuse\_g}
\end{equation}

where $\phi (\cdot)$ represents the normalization and the sigmoid activation applied after the convolution.

\subsection{Fusion Head Module}

To fully leverage the fine-grained spatial details provided by low-level features and the rich semantic context encoded in high-level features for complementary advantages, the primary task of the fusion head is to aggregate features from multiple stages. Specifically, for point-level features, the consistent topological structure across stages allows for direct feature-level concatenation, enabling the integration of multi-stage information while preserving local details and semantic cues. However, due to the downsampling operations in the backbone network, image features from different stages exhibit varying spatial resolutions. To address this issue, all image features are resized to a consistent spatial resolution using linear interpolation to achieve spatial alignment, denoted as $\text{BilInterp}(\cdot)$:

\begin{equation}
\left\{
\begin{aligned}
    \mathcal{F}_{p}^{out} &= \text{MLPs}([\mathcal{F}_{p}^{1};...;\mathcal{F}_{p}^{S}]) \\
   \hat{\mathcal{F}}_{g}^{i} &= \text{BilInterp}(\mathcal{F}_{g}^{i}, h, w), \quad i = 1, \dots, S \\
    \mathcal{F}_{g}^{out} &= \sigma(\text{Conv}([\hat{\mathcal{F}_{g}^{1}};...;\hat{\mathcal{F}_{g}^{S}}]))
\end{aligned}
\right.
\end{equation}

where $S$ denotes the number of network layers, and $(h,w)$ represents the target resolution obtained through linear interpolation. We observe that the point-level feature $\mathcal{F}_{p}^{out}$ mainly contains local spatial details for describing fine structural characteristics, while the group-level feature $\mathcal{F}_{g}^{out}$ integrates semantic information from a broader receptive field to represent the overall scene characteristics. 

\begin{equation}
\mathcal{F}_{logit} = \text{MLP}(\text{MLP}(\text{InFlat}(\mathcal{F}_{g}^{out}))+\mathcal{F}_{p}^{out}) + \mathcal{F}_{p}^{0}
\end{equation}

Here, $\mathcal{F}_{logit}$ is used to generate the final semantic scores with a linear head for the point over the entire point cloud. 

\begin{table*}[!b]
\caption{The class-wise IoU scores of different LiDAR semantic segmentation approaches on the SemanticKITTI~\cite{behley2019semantickitti} val set. All mIoU scores are given in percentage (\%). The best and second best scores for each class are highlighted in bold and underline.}
\vspace{-0.1cm}


\begin{adjustbox}{width=\linewidth}
\begin{tabular}{rcccccccccccccccccccc}
\toprule
\textbf{Method} & {\textbf{mIoU}
} & {\textbf{car}} &{\textbf{bi.cle}} & {\textbf{mt.cle}} & {\textbf{truck}} & {\textbf{oth.v.}} & {\textbf{pers.}} & {\textbf{b.clst}} & {\textbf{m.clst}} & {\textbf{road}} & {\textbf{park.}} &{\textbf{sidew.}} & {\textbf{oth.g}} & {\textbf{build.}} & {\textbf{fence}} & {\textbf{veget.}} & {\textbf{trunk}} &{\textbf{terra.}} & {\textbf{pole}} & {\textbf{traff.}}
\\
\midrule
RandLA-Net~\cite{hu2020randla} & $50.0$ & $92.0$ & $8.0$ & $12.8$ & $74.8$ & $46.7$ & $52.3$ & $46.0$ & $0.0$ & $93.4$ & $32.7$ & $73.4$ & $0.1$ & $84.0$ & $43.5$ & $83.7$ & $57.3$ & $73.1$ & $48.0$ & $27.0$
\\
RangeNet++ \cite{milioto2019rangenet++} & $51.2$ & $89.4$ & $26.5$ & $48.4$ & $33.9$ & $26.7$ & $54.8$ & $69.4$ & $0.0$ & $92.9$ & $37.0$ & $69.9$ & $0.0$ & $83.4$ & $51.0$ & $83.3$ & $54.0$ & $68.1$ & $49.8$ & $34.0$
\\
SqueezeSegV2~\cite{wu2019squeezesegv2} & $40.8$ & $82.7$ & $15.1$ & $22.7$ & $25.6$ & $26.9$ & $22.9$ & $44.5$ & $0.0$ & $92.7$ & $39.7$ & $70.7$ & $0.1$ & $71.6$ & $37.0$ & $74.6$ & $35.8$ & $68.1$ & $21.8$ & $22.2$
\\
SqueezeSegV3~\cite{xu2020squeezesegv3} & $53.3$ & $87.1$ & $34.3$ & $48.6$ & $47.5$ & $47.1$ & $58.1$ & $53.8$ & $0.0$ & $95.3$ & $43.1$ & $78.2$ & $0.3$ & $78.9$ & $53.2$ & $82.3$ & $55.5$ & $70.4$ & $46.3$ & $33.2$
\\
SalsaNet~\cite{aksoy2020salsanet} & $59.4$ &$90.5$ & $44.6$ & $49.6$ & $86.3$ & $54.6$ & $74.0$ & $81.4$ & $0.0$ & $93.4$ & $40.6$ & $69.1$ & $0.0$ & $84.6$ & $53.0$ & $83.6$ & $64.3$ & $64.2$ & $54.4$ & $39.8$
\\
MinkowskiNet~\cite{MinkNet} & $58.5$ & $95.0$ & $23.9$ & $50.4$ & $55.3$ & $45.9$ & $65.6$ & $82.2$ & $0.0$ & $94.3$ & $43.7$ & $76.4$ & $0.0$ & $87.9$ & $57.6$ & $87.4$ & $67.7$ & $71.5$ & $63.5$ & $43.6$
\\
SPVNAS~\cite{SPVNAS-V1} & $62.1$ & $96.5$ & $44.8$ & $63.1$ & $55.9$ & $64.3$ & $72.0$ & $86.0$ & $0.0$ & $93.9$ & $42.4$ & $75.9$ & $0.0$ & $88.8$ & $59.1$ & $88.0$ & $67.5$ & $73.0$ & $63.5$ & $44.3$
\\
Cylinder3D~\cite{zhou2020cylinder3d} & $65.1$ & $96.4$ & $\mathbf{61.5}$ & $78.2$ & $66.3$ & $69.8$ & $\underline{80.8}$ & $\mathbf{93.3}$ & $0.0$ & $94.9$ & $41.5$ & $78.0$ & $1.4$ & $87.5$ & $55.0$ & $86.7$ & $72.2$ & $68.8$ & $63.0$ & $42.1$
\\
PMF~\cite{zhuang2021perception} & $63.9$ & $95.4$ & $47.8$ & $62.9$ & $68.4$ & $\underline{75.2}$ & $78.9$ & $71.6$ & $0.0$ & $\mathbf{96.4}$ & $43.5$ & $80.5$ & $1.0$ & $88.7$ & $60.1$ & $\underline{88.6}$ & $\underline{72.7}$ & $\underline{75.3}$ & $65.5$ & $43.0$
\\
RangeVit~\cite{ando2023rangevit} & $60.9$ & $94.7$ & $44.1$ & $61.4$ & $71.9$ & $37.7$ & $65.3$ & $75.5$ & $0.0$ & $95.5$ & $48.4$ & $83.1$ & $0.0$ & $88.3$ & $60.0$ & $86.3$ & $65.3$ & $72.7$ & $63.1$ & $42.7$
\\
CENet~\cite{cheng2022cenet} & $60.6$ & $91.6$ & $42.4$ & $61.7$ & $82.4$ & $63.5$ & $64.4$ & $76.6$ & $0.0$ & $93.0$ & $50.3$ & $72.7$ & $0.1$ & $85.0$ & $54.4$ & $84.1$ & $61.0$ & $70.3$ & $55.2$ & $42.8$
\\
MSF-CSCNet~\cite{MSF-CSCNet} & $65.8$ & $96.8$ & $50.9$ & $72.7$ & $\underline{90.4}$ & $61.7$ & $76.2$ & $90.2$ & $0.0$ & $94.3$ & $44.8$ & $81.3$ & $0.5$ & $90.3$ & $59.2$ & $86.9$ & $66.5$ & $71.3$ & $65.5$ & $51.1$
\\
RangeFormer~\cite{kong2023rethinking} & $67.1$ & $95.0$ & $\underline{58.1}$ & $72.1$ & $85.1$ & $59.8$ & $76.9$ & $86.4$ & $\underline{0.2}$ & $94.8$ & $\underline{55.5}$ & $81.7$ & $13.0$ & $88.5$ & $64.5$ & $86.5$ & $66.8$ & $73.0$ & $64.0$ & $52.0$
\\
SphereFormer~\cite{lai2023spherical} & $67.8$ & $96.8$ & $50.7$ & $75.1$ & $\mathbf{93.1}$ & $65.7$ & $76.9$ & $\underline{92.0}$ & $0.0$ & $94.7$ & $53.2$ & $82.2$ & $3.2$ & $90.7$ & $58.4$ & $\mathbf{88.7}$ & $71.2$ & $\mathbf{75.9}$ & $64.6$ & $\mathbf{54.5}$
\\
FRNet~\cite{xu2023frnet} & $67.6$ & $\mathbf{97.2}$ & $53.3$ & $72.9$ & $81.5$ & $72.9$ & $77.2$ & $90.8$ & $\underline{0.2}$ & $95.9$ & $53.7$ & $\mathbf{83.9}$ & $9.0$ & $90.5$ & $65.9$ & $87.0$ & $66.8$ & $72.6$ & $64.0$ & $47.9$
\\
WaffleIron~\cite{puy2023using} & $68.0$ & $96.1$ & $\underline{58.1}$ & $\underline{79.7}$ & $77.4$ & $59.0$ & $\mathbf{81.1}$ & $92.2$ & $\mathbf{1.3}$ & $95.5$ & $50.2$ & $\underline{83.6}$ & $6.0$ & $\mathbf{92.1}$ & $\underline{67.5}$ & $87.8$ & $\mathbf{73.8}$ & $73.0$ & $\underline{65.7}$ & $\underline{52.2}$
\\
FARVNet~\cite{FARVNet} & $\underline{68.5}$ & $\underline{97.0}$ & $54.2$ & $75.9$ & $89.6$ & $72.6$ & $76.0$ & $90.1$ & $0.0$ & $95.7$ & $\mathbf{56.9}$ & $83.4$ & $\mathbf{22.7}$ & $89.8$ & $62.3$ & $87.0$ & $65.9$ & $73.0$ & $63.2$ & $47.1$
\\

DAGLFNet & $\mathbf{69.1}$ & $\underline{97.0}$ & $56.5$ & $\mathbf{82.0}$ & $78.8$ & $\mathbf{75.8}$ & $\mathbf{81.1}$ & $91.6$ & $0.0$ & $\underline{96.0}$ & $47.9$ & $\underline{83.6}$ & $\underline{17.7}$ & $\underline{91.7}$ & $\mathbf{67.6}$ & $87.4$ & $69.3$ & $72.9$ & $\mathbf{66.9}$ & $50.2$

\\\bottomrule
\end{tabular}
\end{adjustbox}
\label{table:semantickitti-class}
\end{table*}

\subsection{Loss Function}
\label{Loss Function}

The final point-wise predictions are supervised by cross-entropy loss to ensure accurate semantic classification of individual points. Since multiple points are aggregated within each frustum, point-level supervision alone may provide insufficient constraints on the intermediate frustum features. We therefore introduce four auxiliary heads and assign each frustum the dominant semantic class as its pseudo-label. Cross-entropy loss provides class-level supervision, Lovász loss directly optimizes the IoU, and boundary loss improves predictions near semantic boundaries. The overall objective is defined as:

\begin{equation}
\mathcal{L}=\mathcal{L}_{\mathrm{point}}+
\sum_{s=1}^{4}\left(
\mathcal{L}_{\mathrm{CE}}^{s}
+1.5\mathcal{L}_{\mathrm{Lovasz}}^{s}
+\mathcal{L}_{\mathrm{boundary}}^{s}
\right)
\end{equation}

\subsection{Data Augmentation}
\label{Data Augmentation}

We apply horizontal and vertical BEV flipping, global rotation within $[-\pi,\pi]$, scaling within $[0.95,1.05]$, and Gaussian translation with a standard deviation of 0.1m along each axis. FrustumMix partitions each projected scan into sensor-specific frustum regions and exchanges them between paired scans, while InstanceCopy transfers selected object instances together with their point-level annotations~\cite{xu2023frnet}.

\section{Experiments}
This section evaluates the efficiency of DAGLFNet. We first describe the benchmark datasets and implementation protocols. Subsequently, quantitative and qualitative experimental results demonstrate the effectiveness of DAGLFNet, highlighting its favorable balance between accuracy and computational efficiency. Comprehensive ablation studies elucidate the contribution of each network component.

\subsection{Datasets}
We conducted comprehensive evaluations on two widely used autonomous driving LiDAR datasets and one indoor 3D semantic segmentation dataset, covering both outdoor driving scenes and indoor environments. SemanticKITTI \cite{behley2019semantickitti}, the dataset was collected using Velodyne HDL-64E sensors in Karlsruhe, Germany. Following the official split, 19130 LiDAR scans are used for training and 4071 scans for validation. Each scene contains approximately 120,000 points annotated across 28 semantic categories. The vertical field of view (FoV) spans $-25^\circ$ to $3^\circ$. nuScenes \cite{Panoptic-nuScenes}, the dataset was captured using a 32-beam LiDAR and contains 28130 training samples and 6019 validation samples. It includes 1,000 driving scenes with densely annotated point clouds across 32 classes, of which 16 categories are evaluated for semantic segmentation. The vertical FoV ranges from $-30^\circ$ to $10^\circ$. ScanNet \cite{dai2017scannet} is a large-scale indoor benchmark consisting of RGB-D scans with dense 3D reconstructions and point-wise semantic annotations. Following the standard split, 1,201 scenes are used for training and 312 scenes for validation.

\subsection{Evaluation Metrics}

Segmentation performance was evaluated using the mIoU and mean Accuracy (mAcc). 
For selected semantic classes, the IoU measures the ratio of correctly predicted points to the total number of points that either belong to or are predicted as that class, 
while the accuracy represents the proportion of correctly classified points within the class.

\begin{table*}[!htbp]
\caption{The class-wise IoU scores of different LiDAR semantic segmentation approaches on the val set of nuScenes~\cite{Panoptic-nuScenes}. All IoU scores are given in percentage (\%). The best and second best scores for each class are highlighted in bold and underline.}
\vspace{-0.1cm}

\begin{adjustbox}{width=\linewidth}
\begin{tabular}{rccccccccccccccccc}
\toprule

\textbf{Method} &
\textbf{mIoU} &
\textbf{bar.} &
\textbf{bi.cle} &
\textbf{bus} &
\textbf{car} &
\textbf{con.v.} &
\textbf{mt.cle} &
\textbf{ped.} &
\textbf{t.cone} &
\textbf{trail.} &
\textbf{truck} &
\textbf{dr.s.} &
\textbf{oth.f.} &
\textbf{sidew.} &
\textbf{terra.} &
\textbf{man.} &
\textbf{veget.}
\\\midrule
AF2S3Net~\cite{cheng20212} & $62.2$ & $60.3$ & $12.6$ & $82.3$ & $80.0$ & $20.1$ & $62.0$ & $59.0$ & $49.0$ & $42.2$ & $67.4$ & $94.2$ & $68.0$ & $64.1$ & $68.6$ & $82.9$ & $82.4$
\\
RangeNet++ \cite{milioto2019rangenet++} & $65.5$ & $66.0$ & $21.3$ & $77.2$ & $80.9$ & $30.2$ & $66.8$ & $69.6$ & $52.1$ & $54.2$ & $72.3$ & $94.1$ & $66.6$ & $63.5$ & $70.1$ & $83.1$ & $79.8$
\\
PolarNet~\cite{zhang2020polarnet} & $71.5$ & $74.7$ & $28.2$ & $85.3$ & $90.9$ & $35.1$ & $77.5$ & $71.3$ & $58.8$ & $57.4$ & $76.1$ & $96.5$ & $71.1$ & $74.7$ & $74.0$ & $87.3$ & $85.7$
\\
PCSCNet~\cite{park2023pcscnet} & $72.0$ & $73.3$ & $42.2$ & $87.8$ & $86.1$ & $44.9$ & $82.2$ & $76.1$ & $62.9$ & $49.3$ & $77.3$ & $95.2$ & $66.9$ & $69.5$ & $72.3$ & $83.7$ & $82.5$
\\
SalsaNext~\cite{SalsaNext} & $72.0$ & $74.8$ & $34.1$ & $85.9$ & $88.4$ & $42.2$ & $72.4$ & $72.2$ & $63.1$ & $61.3$ & $76.5$ & $96.0$ & $70.8$ & $71.2$ & $71.5$ & $86.7$ & $84.4$
\\
SVASeg~\cite{zhao2022svaseg} & $74.7$ & $73.1$ & $44.5$ & $88.4$ & $86.6$ & $48.2$ & $80.5$ & $77.7$ & $65.6$ & $57.5$ & $82.1$ & $96.5$ & $70.5$ & $74.7$ & $74.6$ & $87.3$ & $86.9$
\\
RangeViT~\cite{ando2023rangevit} & $75.2$ & $75.5$ & $40.7$ & $88.3$ & $90.1$ & $49.3$ & $79.3$ & $77.2$ & $66.3$ & $65.2$ & $80.0$ & $96.4$ & $71.4$ & $73.8$ & $73.8$ & $89.9$ & $87.2$
\\
MSF-CSCNet~\cite{MSF-CSCNet} & $75.2$ & $75.1$ & $40.3$ & $91.3$ & $88.3$ & $45.8$ & $78.7$ & $78.7$ & $64.5$ & $64.1$ & $80.9$ & $96.5$ & $73.2$ & $75.0$ & $75.6$ & $88.1$ & $87.3$
\\
Cylinder3D~\cite{zhou2020cylinder3d} & $76.1$ & $76.4$ & $40.3$ & $91.2$ & $\mathbf{93.8}$ & $51.3$ & $78.0$ & $78.9$ & $64.9$ & $62.1$ & $84.4$ & $96.8$ & $71.6$ & $76.4$ & $75.4$ & $90.5$ & $87.4$
\\
AMVNet~\cite{liong2020amvnet} & $76.1$ & $\mathbf{79.8}$ & $32.4$ & $82.2$ & $86.4$ & $\mathbf{62.5}$ & $81.9$ & $75.3$ & $\underline{72.3}$ & $\mathbf{83.5}$ & $65.1$ & $\mathbf{97.4}$ & $67.0$ & $\mathbf{78.8}$ & $74.6$ & \underline{$90.8$} & $87.9$
\\
LEST~\cite{LEST} & $77.1$ & $\underline{79.1}$ & $47.4$ & $91.8$ & $87.5$ & $49.2$ & $86.1$ & $\underline{82.4}$ & $70.5$ & $58.6$ & $80.6$ & $96.9$ & $71.4$ & $\underline{76.8}$ & $\mathbf{77.1}$ & $90.2$ & $88.5$
\\
RPVNet~\cite{xu2021rpvnet} & $77.3$ & $78.2$ & $43.4$ & $92.7$ & \underline{$93.2$} & $49.0$ & $85.7$ & $80.5$ & $66.0$ & $66.9$ & $84.0$ & $96.9$ & $73.5$ & $75.9$ & $70.6$ & $90.6$ & \underline{$88.9$}
\\
WaffleIron~\cite{puy2023using} & $77.6$ & $78.7$ & $\underline{51.3}$ & $93.6$ & $88.2$ & $47.2$ & $86.5$ & $81.7$ & $68.9$ & $69.3$ & $83.1$ & $96.9$ & $74.3$ & $75.6$ & $74.2$ & $87.2$ & $85.2$
\\
RangeFormer~\cite{kong2023rethinking} & $78.1$ & $78.0$ & $45.2$ & $\underline{94.0}$ & $92.9$ & $\underline{58.7}$ & $83.9$ & $77.9$ & $69.1$ & $63.7$ & {$\mathbf{85.6}$} & $96.7$ & $74.5$ & $75.1$ & $75.3$ & $89.1$ & $87.5$
\\
SphereFormer~\cite{lai2023spherical} & $\mathbf{78.4}$ & $77.7$ & $43.8$ & $\mathbf{94.5}$ & $93.1$ & $52.4$ & $\mathbf{86.9}$ & $81.2$ & $65.4$ & \underline{$73.4$} & $85.3$ & $\underline{97.0}$ & $73.4$ & $75.4$ & $75.0$ & $\mathbf{91.0}$ & $\mathbf{89.2}$
\\
WaffleAndRange~\cite{fusaro2024exploiting} & $77.6$ & $78.5$ & $49.6$ & $91.8$ & $87.6$ & $52.7$ & $86.7$ & $82.2$ & $70.1$ & $67.2$ & $79.7$ & $\underline{97.0}$ & $74.7$ & $\underline{76.8}$ & $74.9$ & $87.5$ & $85.0$
\\
FRNet~\cite{xu2023frnet} & $76.1$ & $77.2$ & $39.5$ & $93.4$ & $88.6$ & $52.1$ & $81.4$ & $75.1$ & $65.7$ & $66.2$ & $79.7$ & $96.8$ & $\underline{75.3}$ & $75.4$ & $75.9$ & $88.4$ & $ 85.4$
\\
FARVNet~\cite{FARVNet} & $77.8$ & $77.8$ & $42.1$ & $\mathbf{94.5}$ & $91.8$ & $54.9$ & $84.5$ & $77.0$ & $66.7$ & $70.2$ & $\underline{85.4}$ & $\underline{97.0}$ & $74.6$ & $76.4$ & $75.9$ & $89.2$ & $ 87.4$
\\
LIF-Seg~\cite{LIF-Seg} & $\underline{78.2}$ & $76.5$ & $\mathbf{51.4}$ & $91.5$ & $89.2$ & $58.4$ & $\underline{86.6}$ & $\mathbf{82.7}$ & $\mathbf{72.9}$ & $65.5$ & $84.1$ & $96.7$ & $73.2$ & $74.4$ & $73.1$ & $87.5$ & $ 87.6$\\
DAGLFNet & $\mathbf{78.4}$ & $78.5$ & $46.5$ & $89.4$ & $90.7$ & $57.1$ & $85.9$ & $79.3$ & $70.3$ & $69.7$ & $83.1$ & $\underline{97.0}$ & $\mathbf{76.0}$ & $76.1$ & $\underline{76.0}$ & $89.9$ & $ 88.2$
\\\bottomrule

\end{tabular}
\end{adjustbox}

\label{table:nuscenes-class-val}
\end{table*}

\begin{table*}[!htbp]
\caption{The class-wise IoU scores of different LiDAR semantic segmentation approaches on the downsampled ScanNet \cite{dai2017scannet} val set (image resolution: 64×1024). All IoU scores are given in percentage (\%). The best and second best scores for each class are highlighted in bold and underline. }
\vspace{-0.1cm}

\begin{adjustbox}{width=\linewidth}
\begin{tabular}{rccccccccccccccccccccc}
\toprule
\textbf{Method} &
\textbf{mIoU} &
\textbf{wall} &
\textbf{floor} &
\textbf{cab.} &
\textbf{bed} &
\textbf{chair} &
\textbf{sofa} &
\textbf{table} &
\textbf{door} &
\textbf{wind.} &
\textbf{b.shelf} &
\textbf{pict.} &
\textbf{count.} &
\textbf{desk} &
\textbf{curt.} &
\textbf{fridge} &
\textbf{s.curt} &
\textbf{toilet} &
\textbf{sink} &
\textbf{b.tub} &
\textbf{o.furn}
\\\midrule

FRNet \cite{xu2023frnet} &$67.1$& $\underline{83.3}$ & $\mathbf{98.1}$ & $56.7$& $78.3$& $77.8$& $69.5$& $82.8$& $42.6$& $\mathbf{59.9}$& $61.3$& $\mathbf{44.1}$& $\underline{57.5}$& $53.0$& $65.2$& $65.0$& $54.5$& $\underline{92.8}$& $65.6$& $49.4$& $\mathbf{83.7}$ \\
FARVNet \cite{FARVNet} &$\underline{68.8}$& $83.1$ & $\mathbf{98.1}$ & $\underline{59.3}$& $\mathbf{78.8}$& $\underline{81.6}$& $\mathbf{74.0}$& $82.8$& $44.0$& $\underline{58.1}$& $\mathbf{62.5}$& $39.7$& $\mathbf{58.3}$& $\underline{56.0}$& $\underline{67.1}$& $\mathbf{75.1}$& $\underline{68.0}$& $92.6$& $\underline{68.4}$& $\underline{49.7}$& $79.3$ \\
DAGLFNet &$\mathbf{69.8}$& $\mathbf{83.8}$ & $\underline{97.7}$ & $\mathbf{60.3}$ & $77.7$& $\mathbf{82.2}$ & $\underline{72.8}$ & $\mathbf{83.8}$&  $\mathbf{47.2}$ & $57.2$& $\underline{62.2}$ & $\underline{43.9}$ & $55.3$& $\mathbf{58.1}$ & $\mathbf{67.6}$ & $\underline{68.9}$ & $\mathbf{70.2}$ & $\mathbf{93.1}$ & $\mathbf{72.5}$ & $\mathbf{58.8}$ & $\underline{82.3}$
\\\bottomrule

\end{tabular}
\end{adjustbox}
\label{table:scannet-class-val}
\end{table*}

\subsection{Implementation Details}

For ScanNet \cite{dai2017scannet}, the original dense point clouds are converted into virtual LiDAR scans at three range-image resolutions: $32\times1024$, $64\times1024$, and $128\times1024$. For each resolution, points are transformed from Cartesian coordinates into spherical coordinates and assigned to the corresponding angular bins. During the construction of the virtual LiDAR scans, if multiple ScanNet points are assigned to the same viewing ray, only the point with the smallest radial distance is retained. Since ScanNet does not provide reflectance intensity, the normalized radial depth is used as a substitute for the intensity channel, thereby maintaining a consistent input representation across datasets. The resulting virtual scans are evaluated separately to assess the model performance under different point densities. Although SemanticKITTI \cite{behley2019semantickitti} and nuScenes \cite{Panoptic-nuScenes} differ in LiDAR beam count (64 vs. 32), both utilize a full $360^{\circ}$ horizontal field of view (FoV). The range image resolutions were set to $64\times1024$ and $32\times1024$ for SemanticKITTI \cite{behley2019semantickitti} and nuScenes \cite{Panoptic-nuScenes}, respectively. Each projected point is represented by its 3D coordinates and reflectance intensity. The feature encoder uses channel dimensions of (64,128,256,256) and compresses the output to 20 channels, while the four backbone stages each output 128 channels. The model is trained for 250,000 iterations using AdamW with an initial learning rate of 0.001, a batch size of 4, and the OneCycle learning-rate policy. In this work, FRNet \cite{xu2023frnet} is adopted as the main baseline for comparison. All experiments were conducted on a single NVIDIA RTX 4090 GPU.

\begin{figure*}[!h]
    \centering
    \includegraphics[width=\linewidth]{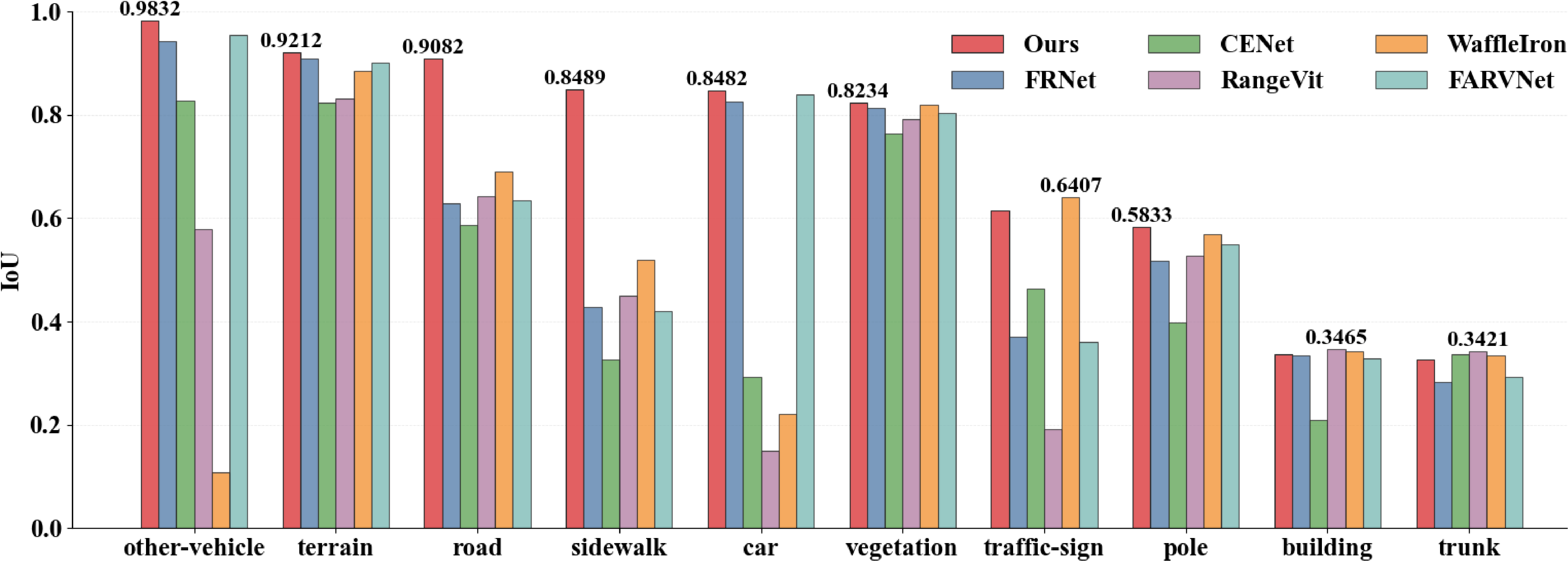}
    \caption{Segmentation IoU results of DAGLFNet and competing methods for points within the 20-60m range on the SemanticKITTI \cite{behley2019semantickitti} val set. }
    \label{comapre_different_range_class}
\end{figure*}

\subsection{Quantitative Results}

We systematically compare DAGLFNet with several state-of-the-art network architectures to comprehensively evaluate performance on the SemanticKITTI~\cite{behley2019semantickitti}, nuScenes~\cite{Panoptic-nuScenes} and ScanNet~\cite{dai2017scannet} datasets. The evaluation encompasses both accuracy and computational efficiency, demonstrating that DAGLFNet achieves an excellent balance between the two and exhibits outstanding overall capability.

The results presented in Tables \ref{table:semantickitti-class} and \ref{table:nuscenes-class-val} demonstrate the competitive performance of DAGLFNet. The results demonstrate that DAGLFNet achieves competitive overall mIoU and leads in several categories. On the SemanticKITTI \cite{behley2019semantickitti} validation set, DAGLFNet achieves an mIoU of $69.1\%$ , surpassing WaffleIron \cite{puy2023using} by $1.1\%$ and FARVNet \cite{FARVNet} by $0.6\%$. On the nuScenes \cite{Panoptic-nuScenes} validation set, it achieves $78.4\%$, outperforming WaffleAndRange \cite{fusaro2024exploiting} by $0.8\%$ and FARVNet \cite{FARVNet} by 0.6\%, ranking among the top performers across multiple categories. Unlike large-scale outdoor scenes, indoor scenes feature smaller spatial scales with more fine-grained semantic category definitions. Table \ref{table:scannet-class-val} shows that DAGLFNet can still be adapted to the smaller-scale indoor dataset ScanNet~\cite{dai2017scannet}. DAGLFNet achieves the best mIoU of 69.8\%, outperforming FRNet \cite{xu2023frnet} and FARVNet \cite{FARVNet} by 2.7\% and 1.0\%, respectively.

The proposed GL-FFE, MB-FE, and FFDFA modules jointly contribute to the improved segmentation performance of DAGLFNet, particularly in distant regions with increasingly sparse point distributions. Fig.~\ref{comapre_different_range_class} presents the class-wise IoU results of DAGLFNet and the competing methods for points within the 20-60m range on the SemanticKITTI \cite{behley2019semantickitti} validation set. DAGLFNet achieves higher IoU scores for most semantic categories, with particularly notable advantages for road and sidewalk, demonstrating its stronger ability to distinguish geometrically similar classes in distant regions. As shown in Table~\ref{tab:range-density}, the regional point density decreases substantially with increasing distance, leading to an overall decline in segmentation performance. DAGLFNet maintains strong segmentation performance in sparse regions, achieving mIoU scores of 44.9\% and 30.0\% at 40-50 m and 50-60 m, respectively, exceeding the second-best results by 1.6 and 4.0 percentage points.

\begin{table}[!htbp]
\centering
\scriptsize
\caption{Distance-wise semantic segmentation performance (mIoU, \%) on SemanticKITTI val set under different regional point densities. Density ($\mathrm{points}/\mathrm{m}^{3}$) is averaged over non-empty 0.5 m cubic voxels across all frames.}
\label{tab:range-density}
\setlength{\tabcolsep}{3pt}
\resizebox{\linewidth}{!}{
\begin{tabular}{l c c c c c c} 
\toprule
\textbf{Method} & \textbf{0-5 m} & \textbf{5-10 m} & \textbf{10-20 m} &\textbf{20-40 m} & \textbf{40-50 m} & \textbf{50-60 m} \\
\textbf{Density} & \textbf{923.8} & \textbf{331.9} & \textbf{88.2} & \textbf{29.8} & \textbf{16.2} & \textbf{13.1} \\
\midrule
SqueezeSegV2~\cite{wu2019squeezesegv2} & 43.6 & 43.1 & 38.3 & 27.7 & 19.5 & 9.3\\
FIDNet~\cite{zhao2021fidnet} & 56.0 & 61.3 &58.5 &48.4 &36.0 &21.1\\
RangeVit~\cite{ando2023rangevit} &58.6 & 62.7 &60.6 & 50.5 & 40.7 & 23.3\\
SegNet4D~\cite{wang2025segnet4d} & 65.4 & 62.8 &61.7 &56.0 &43.3 &26.0\\
Cylinder3D~\cite{zhou2020cylinder3d} & 68.4 & 66.6 & 63.8 &58.7 & 42.2 &15.2\\
WaffleIron~\cite{puy2023using} & 72.8 & 68.7 & 66.1 & 60.3 & 41.7 & 11.6\\
FRNet~\cite{xu2023frnet} & 73.9 & 68.7 &64.8 & 55.6 & 41.5 & 24.2\\
FARVNet~\cite{FARVNet} & 73.7 & 68.4 &65.6 & 57.7 & 40.3 & 22.6\\
DAGLFNet & 74.3 &69.6 & 66.3 &59.1 &44.9 &30.0\\
\bottomrule
\end{tabular}
}
\end{table}


\begin{figure}[!h]
    \centering
    \includegraphics[width=\linewidth]{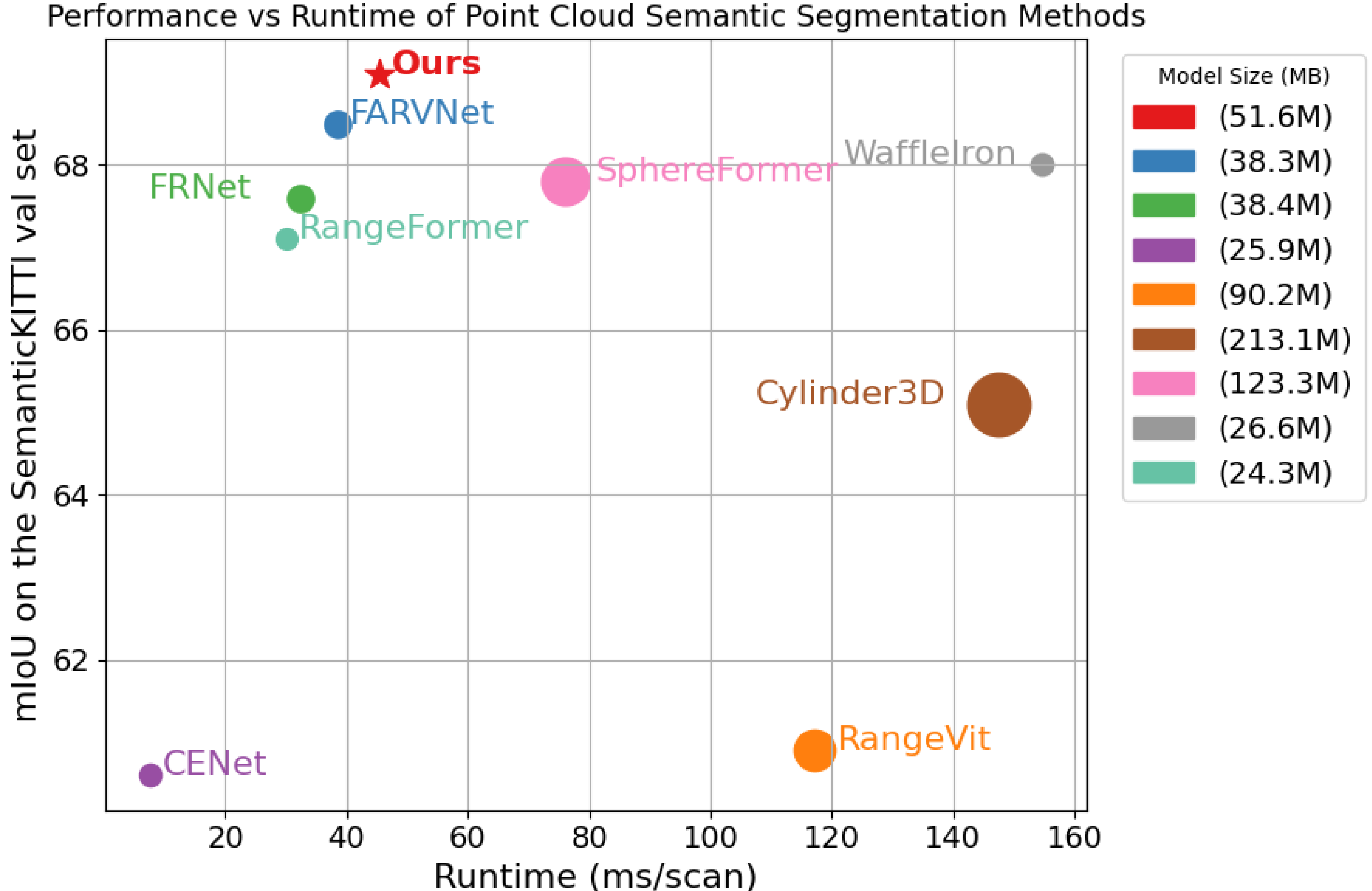}
    \caption{mIoU vs. inference speed for various point cloud semantic segmentation methods on the SemanticKITTI \cite{behley2019semantickitti} validation set. Marker size indicates model size. DAGLFNet achieves the best balance of accuracy and speed.}
    \label{run time}
\end{figure}

Fig.~\ref{run time} illustrates the relationship between mIoU and inference speed (ms/scan) on the SemanticKITTI \cite{behley2019semantickitti} val set for several state-of-the-art point cloud semantic segmentation methods. Our method, DAGLFNet, achieves the highest $69.1\%$ mIoU while maintaining a moderate inference speed of 45.4~ms/scan, striking an optimal balance between accuracy and efficiency. Methods like FRNet \cite{xu2023frnet} and FARVNet \cite{FARVNet} exhibit slightly lower accuracy with comparable inference speeds, whereas faster methods such as CENet \cite{cheng2022cenet} achieve very low latency (7.6~ms/scan) but with significantly lower accuracy ($60.6\%$ mIoU). Similarly, WaffleIron \cite{puy2023using} shows fast inference but slightly reduced accuracy. Overall, DAGLFNet demonstrates a well-balanced trade-off between accuracy, inference efficiency, and model complexity.

\subsection{Qualitative Results}

\begin{figure}[htbp] \centering \begin{adjustbox}{max width=\linewidth} \begin{tikzpicture} \node[inner sep=0pt, anchor=south west] (img) {\includegraphics[width=\linewidth]{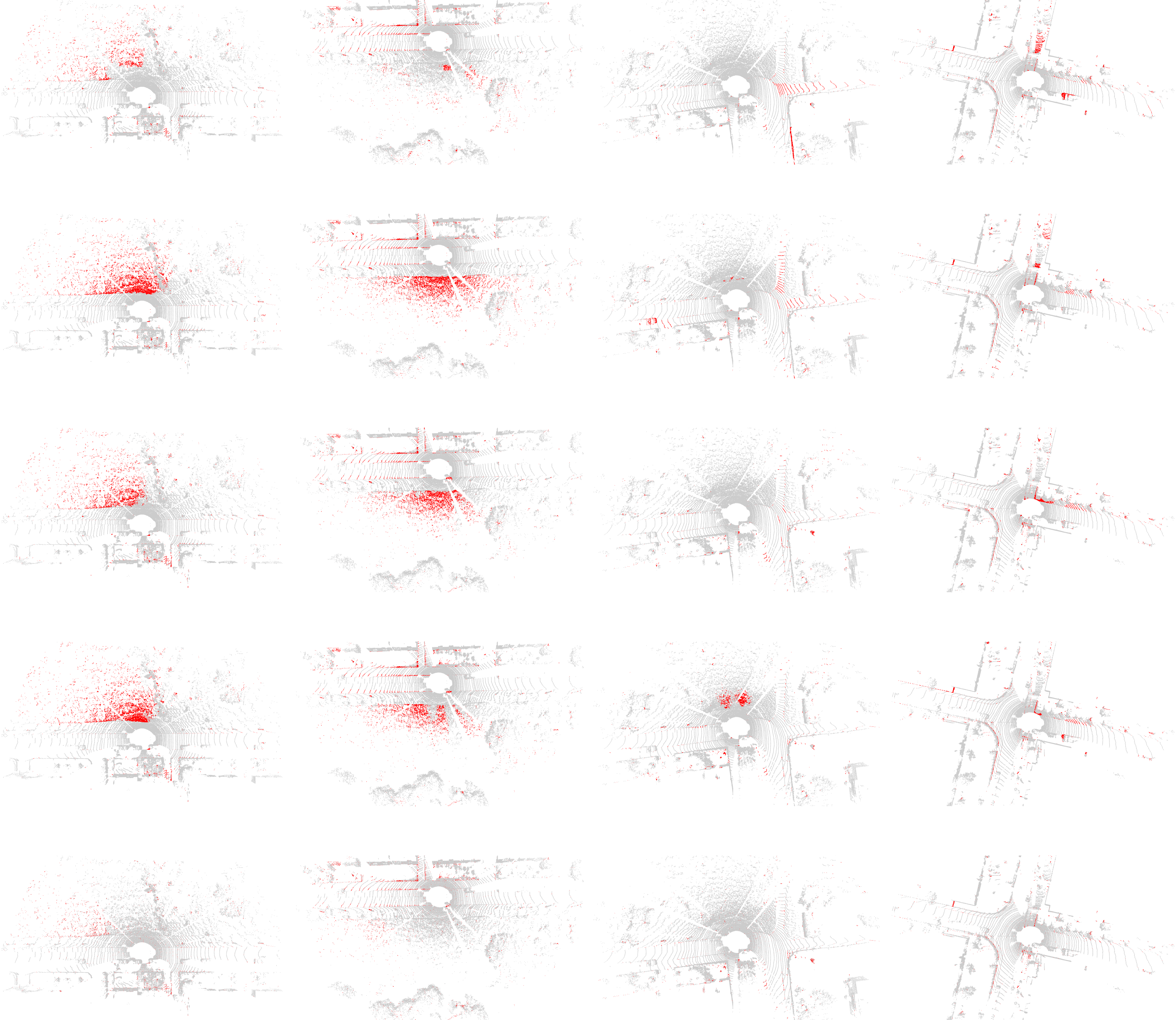}}; \begin{scope}[x={(img.south east)}, y={(img.north west)}] \def\nrows{5} \def\xlabel{-0.05} \def\yoffset{-10/\nrows} \foreach \i/\name/\color in { 1/\textbf{SphereFormer}/cyan!10, 2/\textbf{WaffleIron}/blue!15, 3/\textbf{FARVNet}/green!15, 4/\textbf{FRNet}/orange!15, 5/\textbf{Ours}/purple!15 }{ \pgfmathsetmacro{\ypos}{1 - (\i - 0.5)/\nrows} \node[ rectangle, rounded corners=2pt, fill=\color, minimum width=1.3cm, minimum height={\dimexpr1cm/\nrows}, inner sep=1.5pt, align=center, anchor=center, rotate=90, transform shape, draw=none, font=\scriptsize ] at (\xlabel,\ypos) {\name}; } \def\ncols{4} \foreach \j/\label in { 1/{(a)}, 2/{(b)}, 3/{(c)}, 4/{(d)} }{ \pgfmathsetmacro{\xpos}{(\j - 0.5)/\ncols} \node[anchor=north] at (\xpos, -0.00) {\label}; } \end{scope} \end{tikzpicture} \end{adjustbox} \caption{ Segmentation errors in challenging scenes on the SemanticKITTI \cite{behley2019semantickitti} val set. Red represents misclassified areas, and gray represents correctly classified areas. The four representative scenarios include: (a) segmentation of large-scale vegetation point clouds; (b) classification of vegetation in occluded regions; (c) road recognition in complex intersections; and (d) identification of vehicles and obstacles at intersections. DAGLFNet demonstrates higher accuracy and robustness across all scenarios. } \label{compare_error} \end{figure}

Fig.~\ref{compare_error} visualizes segmentation errors in challenging scenes from the SemanticKITTI \cite{behley2019semantickitti} validation set. We examine four specific scenarios: (a) For large-scale vegetation point clouds, other methods struggle to extract the most discriminative features, leading to extensive segmentation errors. In contrast, DAGLFNet accurately segments most complex regions, outperforming others significantly. (b) Furthermore, in occluded areas of similar scenes, DAGLFNet maintains accurate classification of vegetation within the occluded regions, showcasing its effectiveness in local feature recognition. (c) In complex intersection scenarios\text{-}characterized by numerous and diverse categories\text{-}road recognition is critical. However, methods like SphereFormer \cite{lai2023spherical} and WaffleIron \cite{puy2023using} fail to classify roads correctly, whereas DAGLFNet accurately identifies them, ensuring precise scene understanding. (d) Obstacle recognition, particularly vehicles at intersections, is equally crucial. Other methods perform poorly when identifying vehicles near turning intersections.

\begin{figure}[!h]
\centering
\begin{tikzpicture}
  \node[inner sep=0pt, anchor=south west] (img)
    {\includegraphics[width=0.95\linewidth,trim={0pt 2pt 2pt 0pt},
        clip]{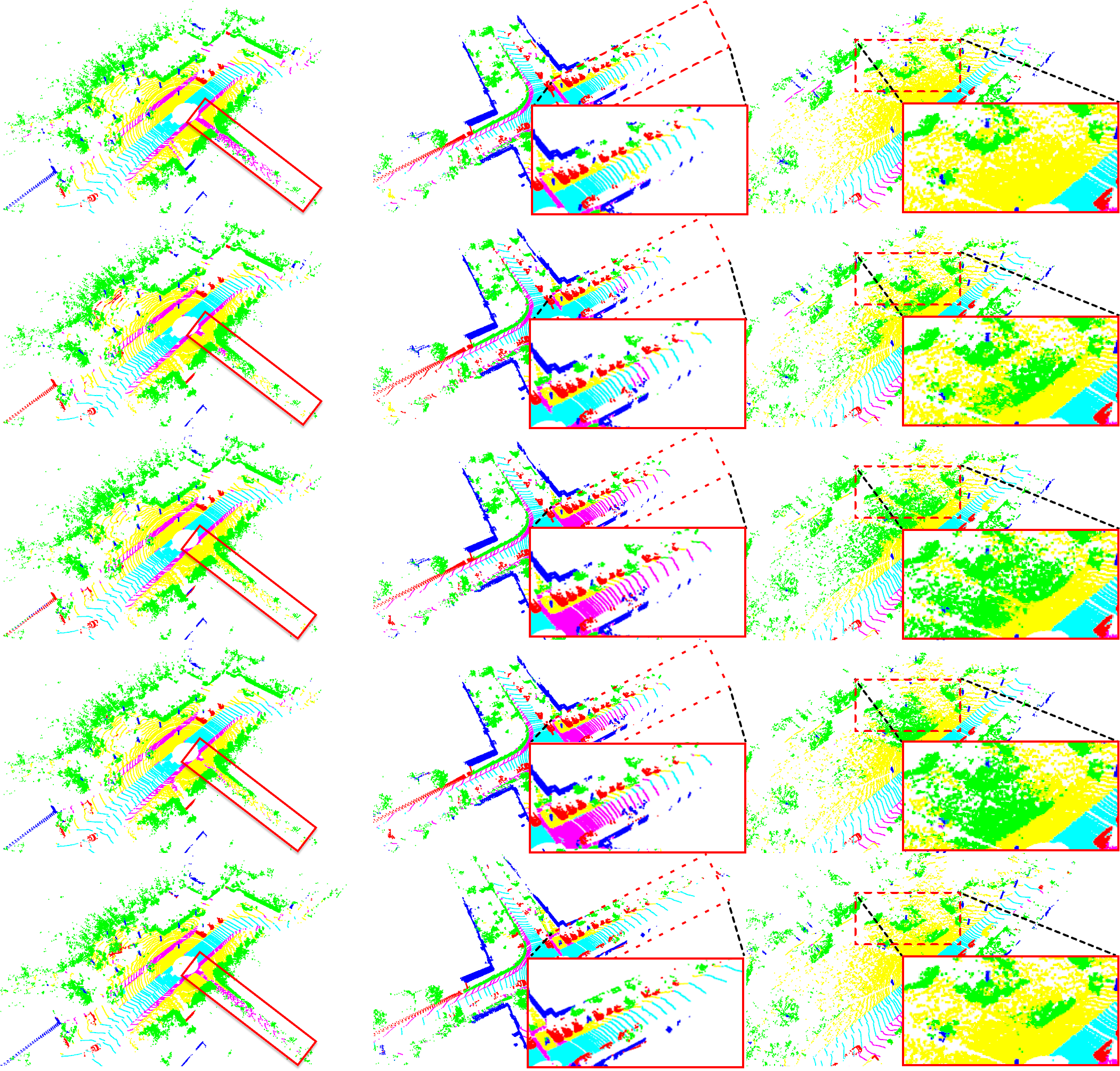}};

  \begin{scope}[x={(img.south east)}, y={(img.north west)}]

    \def\nrows{5} 
    \def\xlabel{-0.02} 
    \def\yoffset{-0.5/\nrows} 
  \foreach \i/\name/\color in {
  1/\textbf{Ground Truth}/cyan!10,
  2/\textbf{waffleiron}/blue!15,
  3/\textbf{FARVNet}/green!15,
  4/\textbf{FRNet}/orange!15,
  5/\textbf{Ours}/purple!15
}{
  \pgfmathsetmacro{\ypos}{1 - (\i - 0.5)/\nrows}
  \node[
    rectangle,
    rounded corners=2pt,
    fill=\color,
    minimum width=1.5cm,  
    minimum height=1/\nrows,  
    inner sep=1.5pt,      
    align=center,
    anchor=center,
    rotate=90,
    transform shape,
    draw=none,
    font=\footnotesize,
  ] at (\xlabel,\ypos) {\name};
}

    \def\ncols{3}
    \foreach \j/\label in {
      1/{(a)},
      2/{(b)},
      3/{(c)}
    }{
      \pgfmathsetmacro{\xpos}{(\j - 0.5)/\ncols}
      \node[anchor=north] at (\xpos, -0.00) {\label};
    }

  \end{scope}
\end{tikzpicture}

\caption{
Three challenging segmentation error scenarios: (a) narrow sidewalks; (b) interference cases; (c) multiple interferences including vegetation, terrain, and occlusions. The regions enclosed by the red boxes are enlarged for clearer visualization.
}
\label{compare_vis}
\end{figure}

To further investigate segmentation errors, we visualize three challenging scenarios in Fig.~\ref{compare_vis}: (a) narrow sidewalks, (b) occlusion cases, and (c) complex interactions involving vegetation, terrain, and occlusions. In these scenes, existing methods exhibit critical limitations: WaffleIron \cite{puy2023using}, FARVNet \cite{FARVNet}, and FRNet \cite{xu2023frnet} misclassify sidewalks as terrain (a); FRNet \cite{xu2023frnet} and FARVNet \cite{FARVNet} incorrectly label occluded roads as sidewalks, while WaffleIron \cite{puy2023using} inconsistently segments overlapping regions (b); and all struggle to distinguish intertwined vegetation from terrain (c). In contrast, DAGLFNet uniquely resolves these challenges, achieving correct segmentation of narrow sidewalks, robust handling of occlusion ambiguities, and precise separation of vegetation-terrain boundaries. These recurring errors may arise because existing methods treat features from different sensing ranges uniformly during projection-based aggregation. As near-range regions contain substantially denser point observations, their responses tend to dominate the fused representation, weakening the relatively sparse geometric cues from distant regions and thereby increasing the risk of misclassification in far-range regions.

\begin{figure}[!h]
    \centering
    \includegraphics[width=\linewidth]{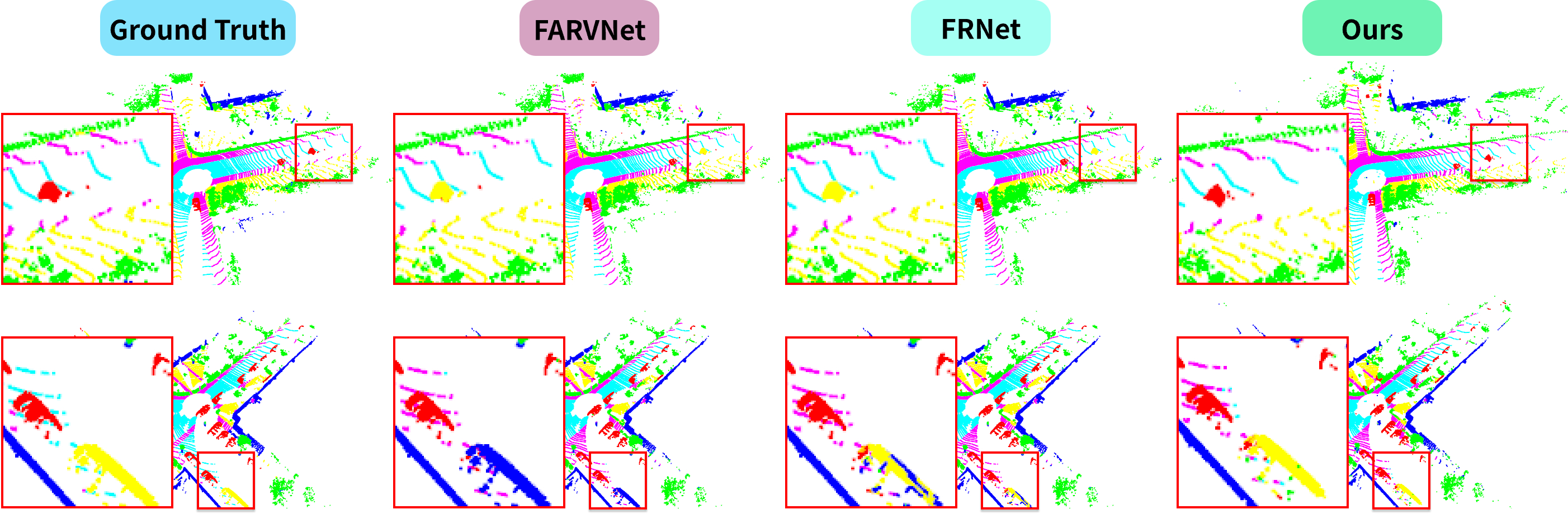}
    \caption{Qualitative comparison of segmentation performance by different methods in distant, sparse, and occluded scenarios. The regions enclosed by the red boxes are enlarged for clearer visualization.}
    \label{compare_depth}
\end{figure}

To further evaluate the performance of DAGLFNet in distant sparse regions, we compared the segmentation of vehicles and buses, as shown in Fig.~\ref{compare_depth}. In distant sparse point clouds, FARVNet \cite{FARVNet} and FRNet \cite{xu2023frnet} tend to confuse sparse vehicles with adjacent sparse terrain, resulting in cars being misclassified as terrain. Under distant occlusion conditions, FARVNet \cite{FARVNet} misclassifies buses as background buildings, while FRNet \cite{xu2023frnet} only partially recognizes them and still confuses them with the background. This is mainly because conventional methods do not fully exploit global context and multi-scale features when processing sparse point clouds and local information. In contrast, DAGLFNet combines local and global features, enhances contextual information, aligns multi-scale features, and integrates a depth-guided attention mechanism, enabling distant and occluded targets to be more effectively distinguished and recognized, thereby achieving higher accuracy.

\begin{figure}[htbp]
    \centering

    \begin{minipage}[b]{0.23\linewidth}
        \centering
        \includegraphics[width=\linewidth]{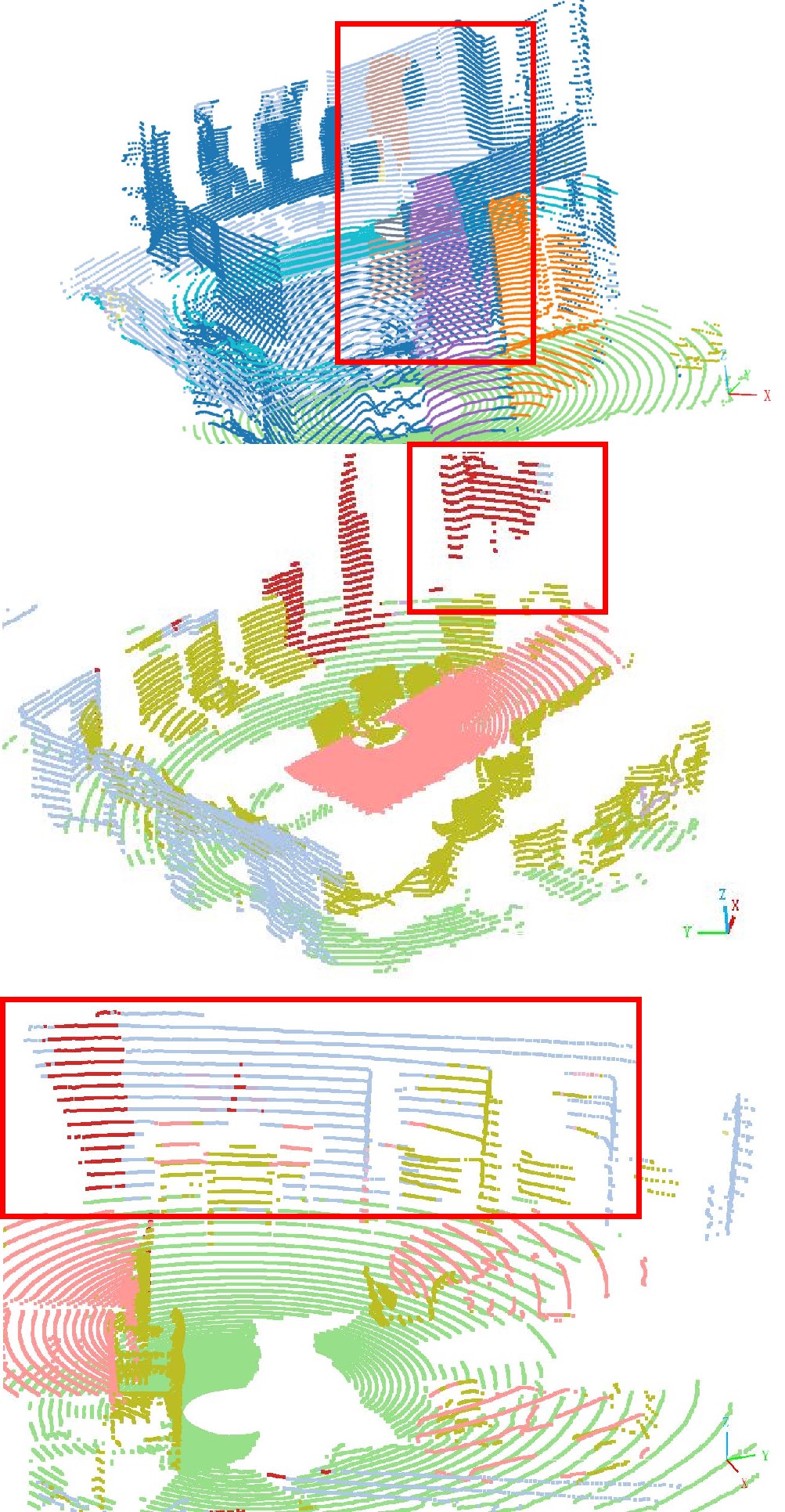}
        
        {\small FRNet}
    \end{minipage}
    \hfill
    \begin{minipage}[b]{0.23\linewidth}
        \centering
        \includegraphics[width=\linewidth]{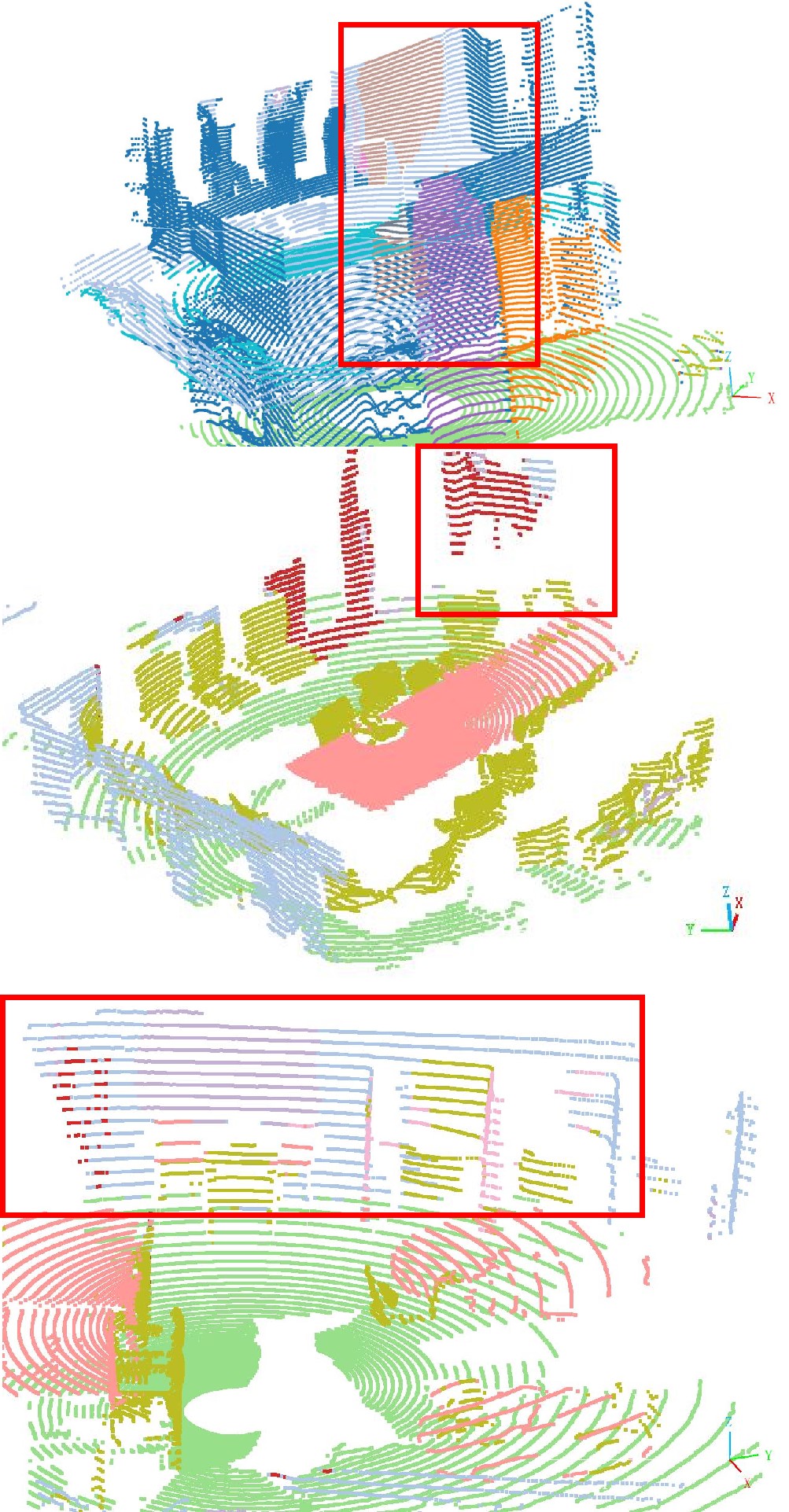}
        
        {\small FARVNet}
    \end{minipage}
    \hfill
    \begin{minipage}[b]{0.23\linewidth}
        \centering
        \includegraphics[width=\linewidth]{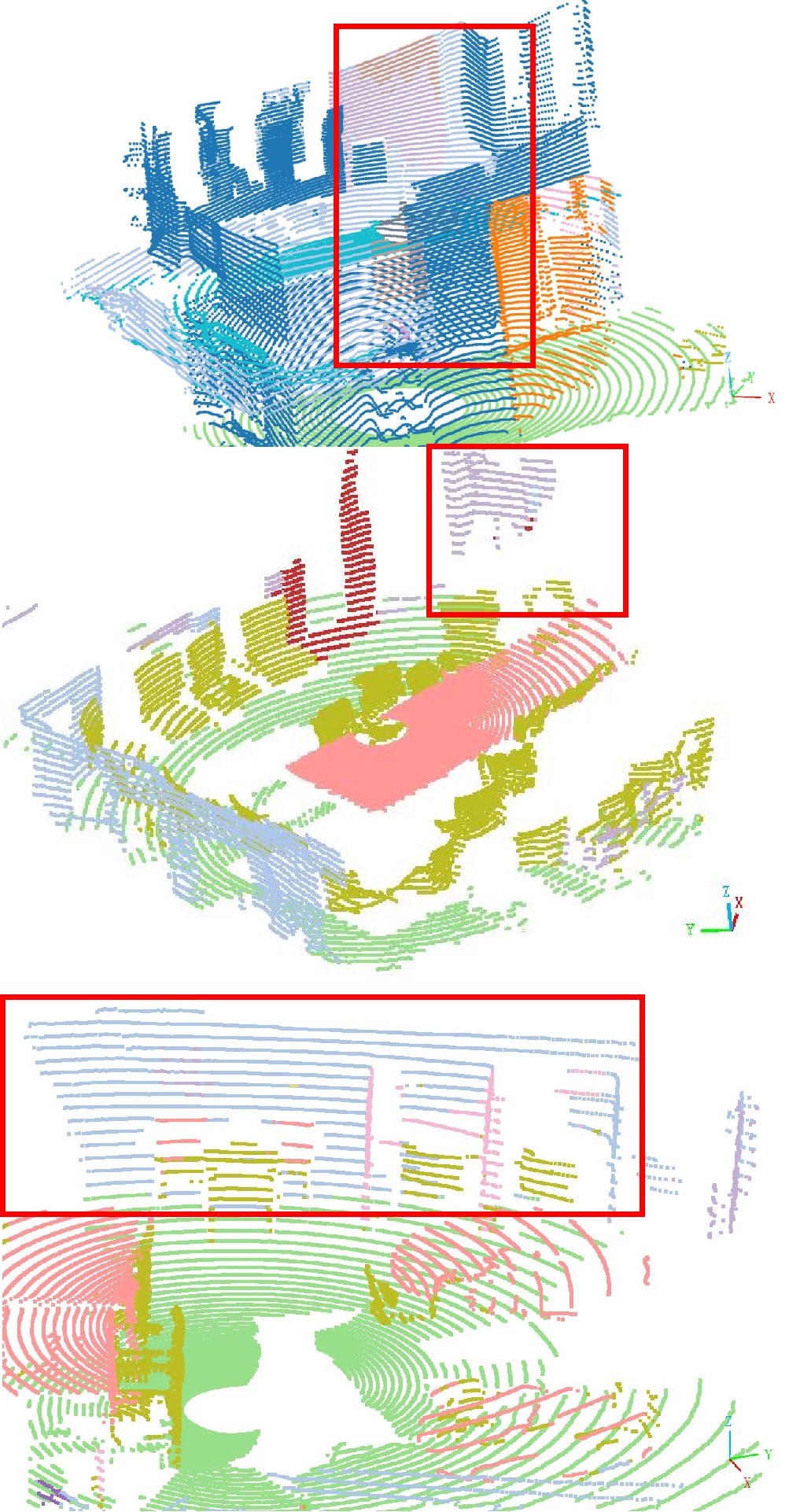}
        
        {\small Ours}
    \end{minipage}
    \hfill
    \begin{minipage}[b]{0.23\linewidth}
        \centering
        \includegraphics[width=\linewidth]{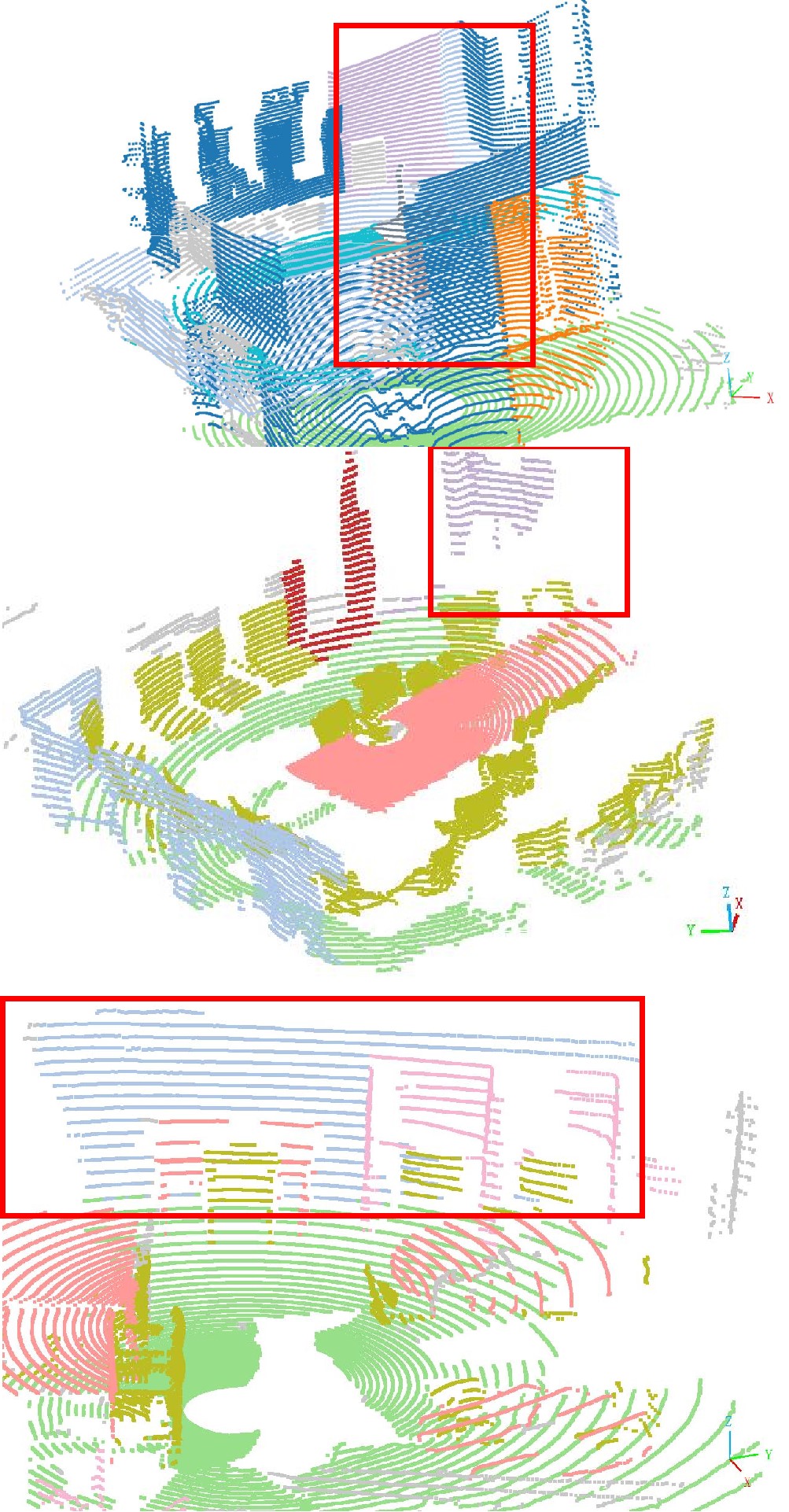}
        
        {\small GT}
    \end{minipage}

    \caption{Qualitative comparison of semantic segmentation results on the ScanNet~\cite{dai2017scannet} dataset. The red boxes highlight regions in which our method provides more accurate predictions and better preserves local structures than the competing methods.}
    \label{fig:comparison-scannet-vis}
\end{figure}

Fig.~\ref{fig:comparison-scannet-vis} further illustrates the segmentation results of the model on the ScanNet~\cite{dai2017scannet} dataset. All models can correctly segment indoor categories such as chairs, tables, and floors. In contrast, FRNet~\cite{xu2023frnet} and FARVNet~\cite{FARVNet} produce less precise boundaries and exhibit intra-class prediction inconsistency. This is due to the denser point distribution and smaller inter-point distances in indoor scenes, where other pseudo-image-based methods fail to learn discriminative features, while DAGLFNet effectively captures such fine-grained variations.

\subsection{Ablation Study}

In this section, we discuss the effectiveness of each design component in the DAGLFNet architecture. All experiments are conducted and reported separately on the validation sets of SemanticKITTI \cite{behley2019semantickitti}, nuScenes \cite{Panoptic-nuScenes} and ScanNet \cite{dai2017scannet}. 

\begin{table}[!htbp]
\centering
\scriptsize
\caption{Ablation study of each component in Ours on the val set of SemanticKITTI \cite{behley2019semantickitti} and nuScenes \cite{Panoptic-nuScenes}. 
BL: Baseline; GL-FFE: Global-Local Feature Fusion Encoding; MB-FE: Multi-Branch Feature Extraction; 
FFDFA: Feature Fusion via Depth Feature-guided Attention; TTA: Test Time Augmentation. 
All mIoU and mAcc scores are given in percentage (\%).}
\label{tab:ablation}
\setlength{\tabcolsep}{3pt}
\resizebox{\linewidth}{!}{
\begin{tabular}{l c c c c | c c | c c} 
\toprule
\multirow{2}{*}{\textbf{BL}} & \multirow{2}{*}{\textbf{GL-FFE}} & \multirow{2}{*}{\textbf{MB-FE}} & \multirow{2}{*}{\textbf{FFDFA}} & \multirow{2}{*}{\textbf{TTA}} & \multicolumn{2}{c}{\textbf{SemKITTI}} & \multicolumn{2}{c}{\textbf{nuScenes}} \\
 & & & & & \textbf{mIoU} & \textbf{mAcc} & \textbf{mIoU} & \textbf{mAcc} \\
\midrule
\checkmark & & & & & 67.6 & 74.0 & 76.1 & 83.9 \\
\checkmark & \checkmark & & & & 67.8 & 74.4 & 76.3 & 84.9 \\
\checkmark & & \checkmark & & & 68.1 & 74.2 & 76.5 & 84.8 \\
\checkmark & & & \checkmark & & 68.5 & 74.6 & 76.9 & 85.0 \\
\checkmark & \checkmark & \checkmark & & & 68.2 & 74.8 & 77.1 & 85.0 \\
\checkmark & \checkmark & \checkmark & \checkmark & & 69.1 & 75.1 & 78.4 & 85.5 \\
\checkmark & \checkmark & \checkmark & \checkmark & \checkmark & 70.0 & 75.5 & 78.7 & 85.7 \\
\bottomrule
\end{tabular}
}
\end{table}

By introducing GL-FFE, which integrates global and local contextual information to enrich point feature representation, the model achieves an improvement of 0.2 mIoU on SemanticKITTI \cite{behley2019semantickitti} and 0.2 mIoU on nuScenes \cite{Panoptic-nuScenes}. This indicates that capturing both global scene context and local details helps to better distinguish complex structures in sparse point clouds. Subsequently, adding MB-FE, which extracts multi-scale and diverse semantic features through parallel branches, further increases mIoU by 0.4 on SemanticKITTI \cite{behley2019semantickitti} and 0.8 on nuScenes \cite{Panoptic-nuScenes}, suggesting that the multi-branch design enables more comprehensive feature extraction across varying object scales. Finally, the FFDFA module, leveraging attention guided by depth features to selectively enhance informative points while suppressing background noise, contributes an additional 0.9 mIoU on SemanticKITTI \cite{behley2019semantickitti} and 1.3 mIoU on nuScenes \cite{Panoptic-nuScenes}, demonstrating its effectiveness in emphasizing critical features and improving segmentation accuracy in challenging regions. Finally, adopting test time augmentation during inference, following prior works, mproves mIoU by 0.9 and 0.3 percentage points, respectively.

\begin{figure}[!h]
    \centering
    \includegraphics[
        width=\linewidth,
        trim={0pt 2pt 0pt 0pt},
        clip
    ]{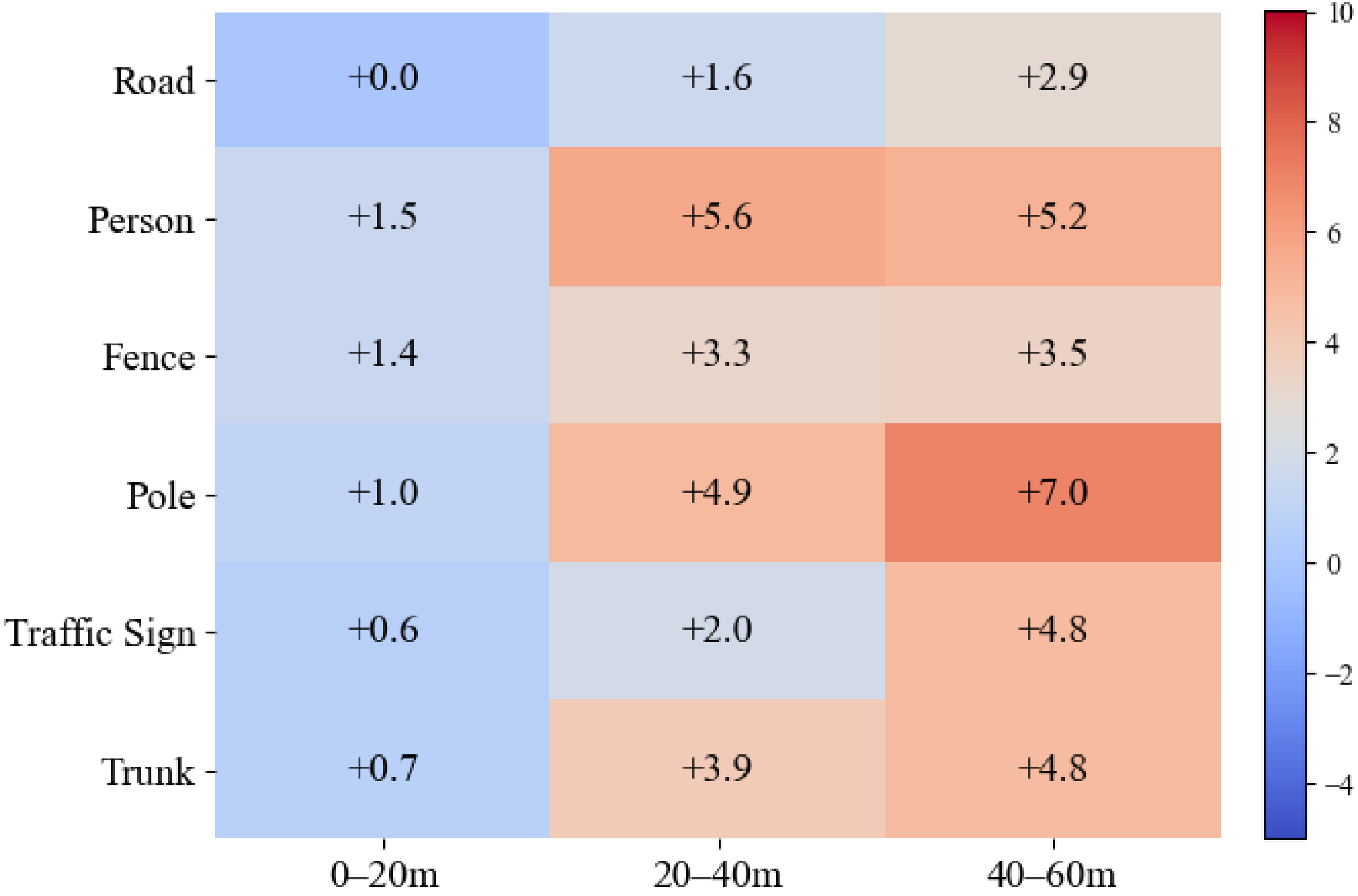}
    \caption{Class-wise IoU gains of DAGLFNet over the baseline at different distance ranges.}
    \label{compare_different_range_IoU_baseline}
\end{figure}

\begin{table*}[!ht]
\caption{Ablation of range image resolution: Evaluation of the impact of different range image resolutions on model performance using the SemanticKITTI~\cite{behley2019semantickitti} val set.}
\vspace{-0.1 cm}

\begin{adjustbox}{width=\linewidth}
\begin{tabular}{rccccccccccccccccccccc}
\toprule
 \textbf{Resolution} & {\textbf{mIoU}
} &{\textbf{FPS}} & {\textbf{car}} &{\textbf{bi.cle}} & {\textbf{mt.cle}} & {\textbf{truck}} & {\textbf{oth.v.}} & {\textbf{pers.}} & {\textbf{b.clst}} & {\textbf{m.clst}} & {\textbf{road}} & {\textbf{park.}} &{\textbf{sidew.}} & {\textbf{oth.g}} & {\textbf{build.}} & {\textbf{fence}} & {\textbf{veget.}} & {\textbf{trunk}} &{\textbf{terra.}} & {\textbf{pole}} & {\textbf{traff.}}
\\
\midrule

$64 \times 256$ & $65.8$ & $30$ & $95.1$ & $50.0$ & $67.4$ & $82.7$ & $55.7$ & $65.3$ & $87.6$ & $0.0$ & $95.0$ & $64.7$ & $83.3$ & $24.7$ & $88.1$ & $60.6$ & $88.8$ & $61.2$ & $79.1$ & $55.8$ & $45.7$
\\
$64 \times 512$ & $68.2$ & $26$ & $96.9$ & $54.3$ & $75.7$ & $71.3$ & $77.1$ & $75.1$ & $91.1$ & $0.0$ & $95.6$ & $63.6$ & $84.3$ & $14.3$ & $90.3$ & $66.5$ & $88.1$ & $66.7$ & $75.1$ & $61.6$ & $48.8$
\\
$64 \times 768$ & $68.0$ & $23$ & $96.7$ & $51.7$ & $79.7$ & $92.2$ & $69.5$ & $77.4$ & $85.2$ & $0.0$ & $95.7$ & $47.3$ & $83.5$ & $18.0$ & $90.9$ & $66.3$ & $87.1$ & $61.6$ & $72.6$ & $66.2$ & $50.3$
\\

$64 \times 1024$ & $69.1$ & $22$ & $97.0$ & $56.5$ & $82.0$ & $78.8$ & $75.8$ & $81.1$ & $91.6$ & $0.0$ & $96.0$ & $47.9$ & $83.6$ & $17.7$ & $91.7$ & $67.6$ & $87.4$ & $69.3$ & $72.9$ & $66.9$ & $50.2$
\\

$64 \times 1280$ & $68.6$ & $17$ & $96.2$ & $54.7$ & $77.2$ & $82.2$ & $68.4$ & $79.6$ & $90.0$ & $19.5$ & $94.9$ & $51.6$ & $84.0$ & $0.0$ & $91.3$ & $68.8$ & $87.6$ & $68.3$ & $73.1$ & $64.6$ & $51.7$

\\\bottomrule
\end{tabular}
\end{adjustbox}
\label{table:range_image}
\end{table*}
\unskip

\begin{table*}[!ht]
\caption{Ablation of network depth: Evaluation of different depth configurations on the SemanticKITTI~\cite{behley2019semantickitti} val set. All mIoU scores are reported in percentage (\%).}
\vspace{-0.1cm}

\begin{adjustbox}{width=\linewidth}
\begin{tabular}{rcccccccccccccccccccccc}
\toprule
 \textbf{Depth} & {\textbf{mIoU}
} &{\textbf{FPS}
}&{\textbf{Params}
} & {\textbf{car}} &{\textbf{bi.cle}} & {\textbf{mt.cle}} & {\textbf{truck}} & {\textbf{oth.v.}} & {\textbf{pers.}} & {\textbf{b.clst}} & {\textbf{m.clst}} & {\textbf{road}} & {\textbf{park.}} &{\textbf{sidew.}} & {\textbf{oth.g}} & {\textbf{build.}} & {\textbf{fence}} & {\textbf{veget.}} & {\textbf{trunk}} &{\textbf{terra.}} & {\textbf{pole}} & {\textbf{traff.}}
\\
\midrule
$[1,1,1,1]$ &$66.1$ & $32$ & $28.2M$ & $95.6$ & $51.1$ & $70.9$ & $81.5$ & $62.6$ & $80.0$ & $86.9$ & $0.0$ & $95.2$ & $47.4$ & $83.0$ & $14.0$ & $89.1$ & $58.0$ & $86.8$ & $67.9$ & $71.3$ & $64.4$ & $50.7$ \\

$[2,2,2,2]$ &$66.9$ & $27$ & $36.0M$ & $96.5$ & $54.0$ & $70.0$ & $80.5$ & $64.9$ & $77.6$ & $89.1$ & $4.5$ & $95.8$ & $52.0$ & $83.4$ & $10.4$ & $89.8$ & $64.0$ & $86.0$ & $67.6$ & $69.2$ & $64.7$ & $50.8$ \\

$[3,3,3,3]$ &$68.3$ & $24$ & $43.8M$ & $96.9$ & $54.6$ & $75.3$ & $88.4$ & $72.0$ & $76.2$ & $91.3$ & $0.1$ & $95.7$ & $56.7$ & $83.5$ & $19.0$ & $89.8$ & $62.7$ & $86.9$ & $64.6$ & $72.8$ & $63.1$ & $47.1$ \\

$[3,4,6,3]$ & $69.1$ & $22$ & $51.6M$ & $97.0$ & $56.5$ & $82.0$ & $78.8$ & $75.8$ & $81.1$ & $91.6$ & $0.0$ & $96.0$ & $47.9$ & $83.6$ & $17.7$ & $91.7$ & $67.6$ & $87.4$ & $69.3$ & $72.9$ & $66.9$ & $50.2$

\\\bottomrule
\end{tabular}
\end{adjustbox}
\label{table:network depth}
\end{table*}
\unskip

To evaluate DAGLFNet across distance ranges, we measure IoU gains for selected semantic classes over three spatial intervals (0-20m, 20-40m, and 40-60m), as shown in Fig.~\ref{compare_different_range_IoU_baseline}. Overall, the improvements tend to become more pronounced as distance increases. In the 40--60\,m range, DAGLFNet achieves notable IoU gains for Pole (+7.0\%), Person (+5.2\%), and Traffic Sign (+4.8\%). These results further demonstrate the effectiveness of the proposed method under sparse far-range point-cloud conditions.

\textit{Range Image Representation.} 
We investigate the effect of range image resolution on DAGLFNet performance. As the resolution decreases, the projected range images become coarser, causing a loss of fine-grained details in the point cloud representation. This reduction primarily affects the accurate extraction of local features and reduces segmentation accuracy for small or distant objects. In contrast, higher resolutions preserve more spatial details, enhance feature extraction, and improve segmentation performance, especially in sparse or complex regions. However, excessively high resolutions increase computational costs, while the performance gains remain limited. As shown in Table~\ref{table:range_image}, we compare the results under different range image resolutions, and the configuration of $64 \times 1024$ achieves the best balance between performance and efficiency.

\textit{Network Depth.} To investigate the influence of network depth on model performance, we conduct an ablation study using different depth configurations of the proposed network, denoted as $[1,1,1,1]$, $[2,2,2,2]$, $[3,3,3,3]$, and $[3,4,6,3]$. As shown in Table~\ref{table:network depth}, increasing network depth generally improves segmentation accuracy, while slightly reducing inference speed and increasing the number of parameters. The configuration $[3,4,6,3]$ achieves the best balance between accuracy and computational efficiency, demonstrating the effectiveness of deeper hierarchical feature extraction in our model.

\textit{Point Cloud Density.} To investigate the influence of point cloud density on model performance, we conduct an ablation study using different LiDAR beam configurations, denoted as 32-, 64-, and 128-beam. As shown in Table~\ref{table:sparsity}, the segmentation performance shows some variation across different point cloud density settings, indicating that point cloud density affects segmentation quality to some extent. However, the overall results demonstrate that DAGLFNet can still maintain effective feature learning under varying point cloud density conditions.

\begin{table}[!htbp]
\centering
\scriptsize
\caption{Ablation study on different point-cloud sparsity levels on the ScanNet~\cite{dai2017scannet} validation set. The original point clouds are converted into virtual LiDAR scans with 32, 64, and 128 vertical scan lines, corresponding to range-image resolutions of $32\times1024$, $64\times1024$, and $128\times1024$, respectively. All mIoU and mAcc scores are reported as percentages (\%).}
\label{table:sparsity}
\setlength{\tabcolsep}{6pt}
\resizebox{0.8\linewidth}{!}{
\begin{tabular}{c c | c c}
\toprule
\textbf{LiDAR Scan Lines}& \textbf{Resolution} & \textbf{mIoU} & \textbf{mAcc} \\
\midrule
32& $32\times 1024$ & 66.9 & 89.2 \\
64& $64\times 1024$ & 69.8 &  89.9\\
128& $128\times 1024$  & 68.3  & 89.7 \\
\bottomrule
\end{tabular}
}
\end{table}

\begin{figure}[!htbp]
   \centering
     \captionsetup[subfloat]{font=scriptsize}
   \subfloat[\scriptsize Ground Truth\label{fig:failure_case_gt}]{
       \includegraphics[width=0.45\linewidth]{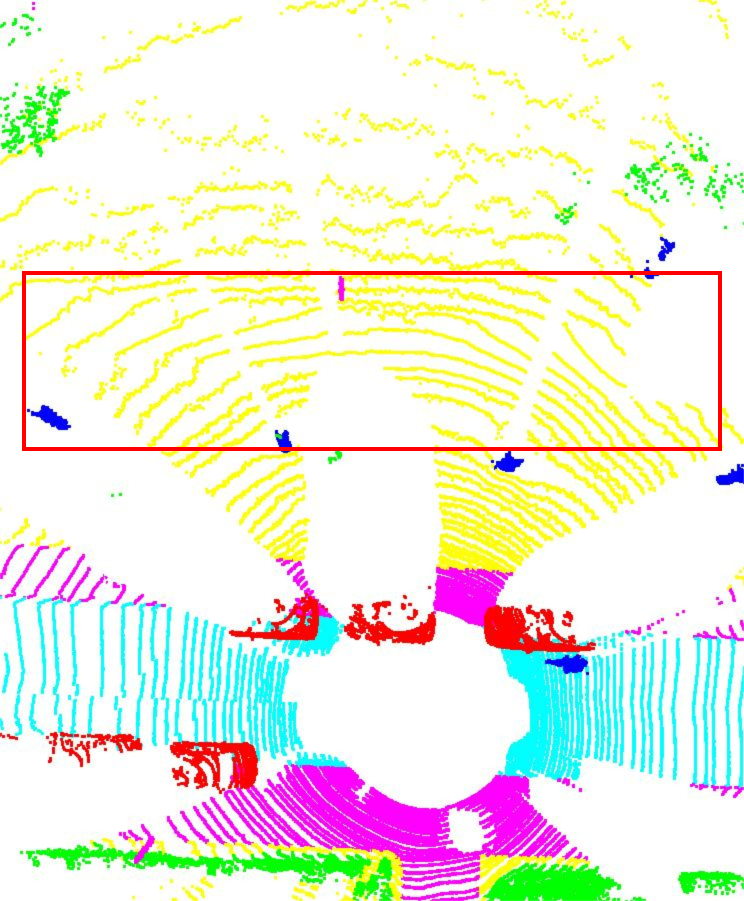}
   }
   \hspace{0.025\linewidth}
   \subfloat[\scriptsize Ours\label{fig:failure_case_error}]{
       \includegraphics[width=0.45\linewidth]{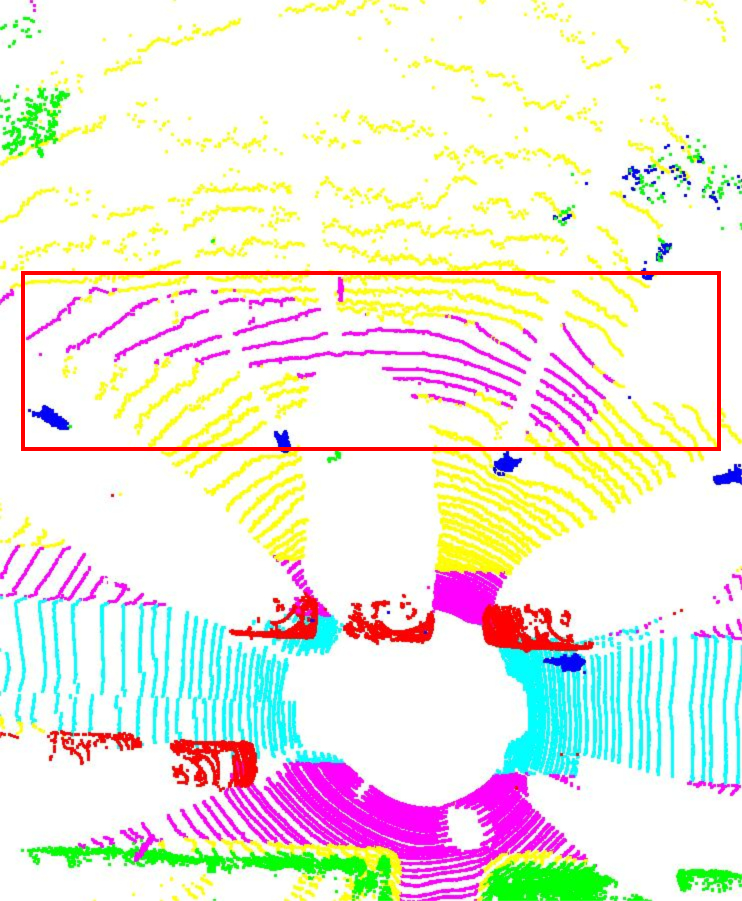}
   }
 \captionsetup[sub]{font=scriptsize}
   \caption{
     Challenging cases involving geometrically similar regions. DAGLFNet may produce inaccurate predictions when ground and sidewalk regions exhibit highly similar local geometric characteristics.}
   
   \label{fig:failure_case}
\end{figure}

\subsection{Limitations}

Variations in terrain, sensor characteristics, and weather conditions may lead to distribution shifts across different data domains, and related studies have also highlighted the challenges posed by such domain variations~\cite{yang2024domain},~\cite{zhang2024learning},~\cite{zhang2024monocular}. In addition, semantically different terrain categories may exhibit highly similar local geometric structures. Although DAGLFNet has achieved promising segmentation performance, it may still produce inaccurate predictions when ground and sidewalk regions share similar local geometric characteristics, as illustrated in Fig.~\ref{fig:failure_case}. Future work will explore multi-domain training to further improve the robustness and generalization ability of DAGLFNet across different data domains.

\section{Conclusion}

In this work, we introduce DAGLFNet, a pseudo-image-based framework for point cloud semantic segmentation, specifically engineered to mitigate challenges arising from point cloud sparsity and occlusions. By fusing local and global features, employing a multi-branch feature extraction mechanism, and utilizing depth feature-guided attention fusion, DAGLFNet achieves robust segmentation performance in far-range, sparse, and complex environments. Extensive results across the SemanticKITTI, nuScenes, and ScanNet benchmarks demonstrate the accuracy and effectiveness of the proposed network in both outdoor and indoor environments.

\bibliographystyle{IEEEtran}
\bibliography{sample}

\newpage

 





\end{document}